\PassOptionsToPackage{table}{xcolor}
\documentclass[sigconf]{acmart}

\AtBeginDocument{%
  }

\copyrightyear{2025}
\acmYear{2025}
\setcopyright{acmlicensed}\acmConference[MM '25]{Proceedings of the 33rd ACM International Conference on Multimedia}{October 27--31, 2025}{Dublin, Ireland}
\acmBooktitle{Proceedings of the 33rd ACM International Conference on Multimedia (MM '25), October 27--31, 2025, Dublin, Ireland}
\acmDOI{10.1145/3746027.3755832}
\acmISBN{979-8-4007-2035-2/2025/10}

\usepackage{algorithm}
\usepackage{algorithmic}
\usepackage{amsmath}
\usepackage{multirow}
\usepackage{booktabs}

\newcommand{\DDN}{CogDDN}

\begin{document}

\title{\DDN: A Cognitive Demand-Driven Navigation with Decision Optimization and Dual-Process Thinking}

\author{Yuehao Huang}
\email{yuehaohuang@zju.edu.cn}
\orcid{0009-0002-9397-8341}
\authornote{Equal contribution}
\affiliation{
  \institution{Zhejiang University}
  \city{Hangzhou}
  \country{China}
  }

\author{Liang Liu}
\orcid{0000-0001-7910-810X}
\authornotemark[1]
\email{liangliu.vivoai@vivo.com}
\affiliation{
  \institution{vivo AI Lab}
  \city{Hangzhou}
  \country{China}
  }

\author{Shuangming Lei}
\email{shuangminglei@zju.edu.cn}
\orcid{0009-0001-5344-6079}
\affiliation{
  \institution{Zhejiang University}
  \city{Hangzhou}
  \country{China}
  }

\author{Yukai Ma}
\orcid{0000-0001-8135-9012}
\email{yukaima@zju.edu.cn}
\affiliation{
  \institution{Zhejiang University}
  \city{Hangzhou}
  \country{China}
  }

\author{Hao Su}
\orcid{0009-0001-5450-1383}
\email{haosu@@zju.edu.com}
\affiliation{
  \institution{Zhejiang University}
  \city{Hangzhou}
  \country{China}
  }

\author{Jianbiao Mei}
\email{jianbiaomei@zju.edu.cn}
\orcid{0000-0003-3849-2736}
\affiliation{
  \institution{Zhejiang University}
  \city{Hangzhou}
  \country{China}
  }

\author{Pengxiang Zhao}
\email{zhaopengxiang@zju.edu.cn}
\orcid{0009-0004-6567-5815}
\affiliation{
  \institution{Zhejiang University}
  \city{Hangzhou}
  \country{China}
  }

\author{Yaqing Gu}
\email{yaqinggu@zju.edu.cn}
\orcid{0009-0003-5918-3235}
\affiliation{
  \institution{Zhejiang University}
  \city{Hangzhou}
  \country{China}
  }

\author{Yong Liu}
\orcid{0000-0003-4822-8939}
\authornote{Corresponding author}
\email{yongliu@iipc.zju.edu.cn}
\affiliation{
  \institution{Zhejiang University}
  \city{Hangzhou}
  \country{China}
  }

\author{Jiajun Lv}
\authornotemark[2]
\orcid{0000-0002-8545-9464}
\email{lvjiajun314@zju.edu.cn}
\affiliation{
  \institution{Zhejiang University}
  \city{Hangzhou}
  \country{China}
  }
  
\renewcommand{\shortauthors}{Yuehao Huang et al.}

\begin{abstract}
Mobile robots are increasingly required to navigate and interact within unknown and unstructured environments to meet human demands. Demand-driven navigation (DDN) enables robots to identify and locate objects based on implicit human intent, even when object locations are unknown. However, traditional data-driven DDN methods rely on pre-collected data for model training and decision-making, limiting their generalization capability in unseen scenarios. In this paper, we propose {\DDN}, a VLM-based framework that emulates the human cognitive and learning mechanisms by integrating fast and slow thinking systems and selectively identifying key objects essential to fulfilling user demands. {\DDN} identifies appropriate target objects by semantically aligning detected objects with the given instructions. Furthermore, it incorporates a dual-process decision-making module, comprising a Heuristic Process for rapid, efficient decisions and an Analytic Process that analyzes past errors, accumulates them in a knowledge base, and continuously improves performance. Chain of Thought (CoT) reasoning strengthens the decision-making process. Extensive closed-loop evaluations on the AI2Thor simulator with the ProcThor dataset show that {\DDN} outperforms single-view camera-only methods by 15\%, demonstrating significant improvements in navigation accuracy and adaptability. The project page is available at \url{https://yuehaohuang.github.io/CogDDN/}.
\end{abstract}


\maketitle

\section{Introduction}

\begin{figure}[t]
  \centering
  \includegraphics[width=0.45\textwidth]{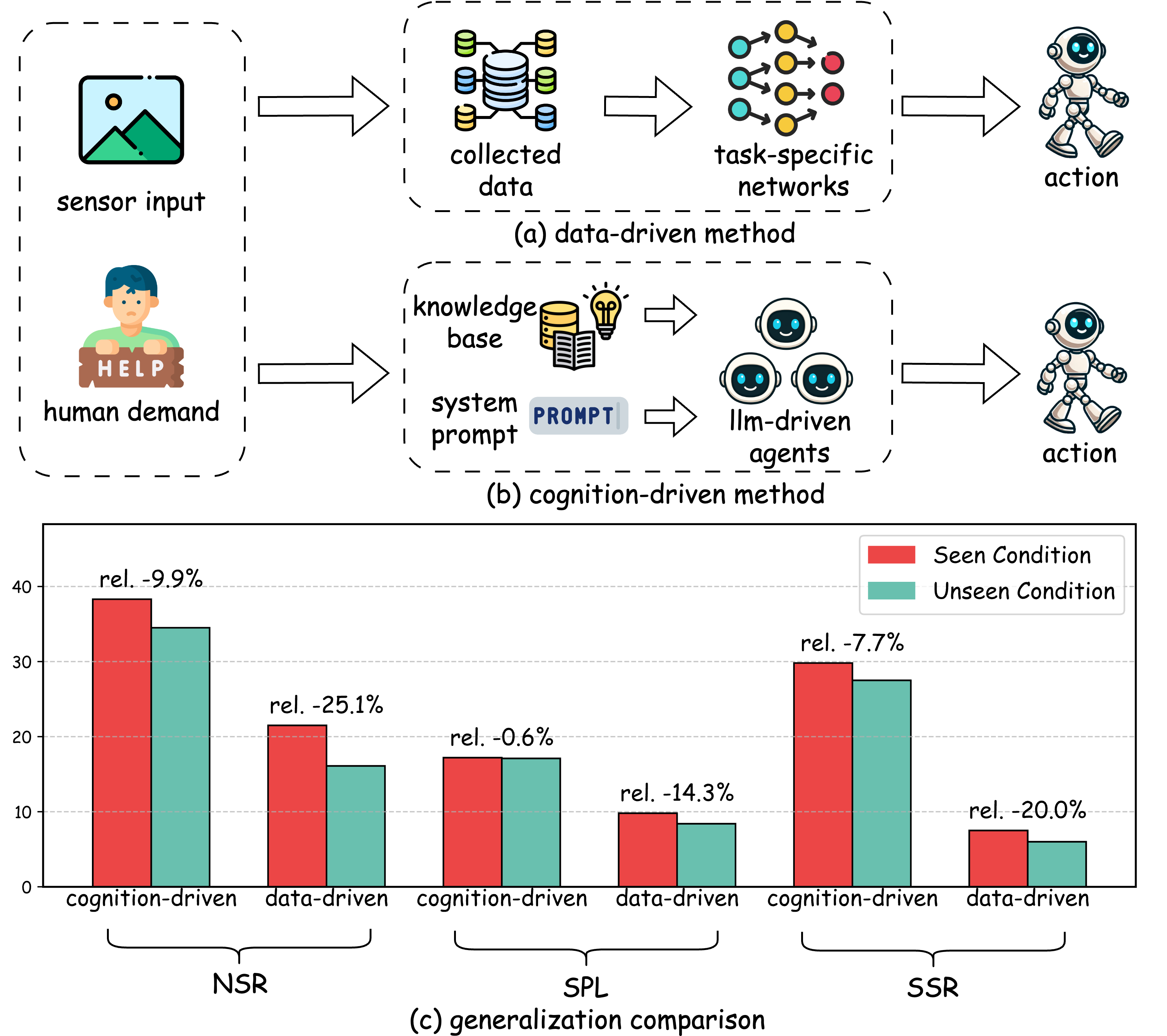}
  \caption{Comparison between cognition-driven and data-driven methods. (a) illustrates the process of the data-driven method, (b) shows the process of the cognition-driven method, and (c) compares the generalization capabilities of both methods using NSR, SPL and SSR metrics, with relative decrease rates marked between seen and unseen conditions.}
  \label{fig:first}
  \Description{Comparison between cognition-driven and data-driven.}
\end{figure}

Mobile robots are becoming indispensable in dynamic and unstructured environments such as households, hospitals, and warehouses. To operate effectively and naturally within these human-centric spaces, they must move beyond simple command execution to interpret and respond to implicit human demands. This capability is particularly critical when target objects are uncertain or not explicitly specified. For instance, a hungry person instinctively seeks food based on internal cues, rather than following a direct instruction like "find the apple on the kitchen counter". Similarly, demand-driven navigation (DDN)~\cite{ddn, mo-ddn} enables robots to locate objects that fulfill such implicit needs without relying on predefined object lists or known locations. However, as illustrated in Figure~\ref{fig:first} (a), conventional data-driven DDN methods~\cite{ddn,mo-ddn} are fundamentally constrained by their heavy reliance on extensive, pre-collected datasets for training and decision-making. This dependence curtails their adaptability and generalization when encountering novel scenarios or vaguely phrased instructions.

Recent breakthroughs in Large Language Models (LLMs) and Vision-Language Models (VLMs) have catalyzed a new era in robotic intelligence, as these models possess remarkable reasoning capabilities and extensive world knowledge that are exceptionally well-suited for complex navigation tasks. Leveraging these advancements, cognition-driven approaches~\cite{navigationvlmframework,WMNav,MapNav,LLM-Nav,llmnav} employ these models to help agents interpret high-level instructions and strategically locate targets. The DDN task, which inherently fuses natural language with real-time visual perception, demands that an agent infer and act upon subtle user intent. As depicted in Figure~\ref{fig:first} (b), our proposed framework synergistically integrates LLMs and VLMs for DDN, enabling sophisticated and nuanced reasoning over ambiguous instructions. The embedded VLMs enhance navigation by jointly processing visual and linguistic inputs, effectively bridging the semantic gap between high-level intent and spatial context. This cognition-driven, multi-agent approach significantly improves the agent's spatial and contextual understanding, yielding more accurate, robust, and adaptable navigation. Figure~\ref{fig:first} (c) shows that the cognition-driven method generalizes better than traditional data-driven approaches, exhibiting a substantially smaller performance drop when transitioning from seen to unseen scenarios.

\begin{figure}[t]
  \centering
  \includegraphics[width=0.45\textwidth]{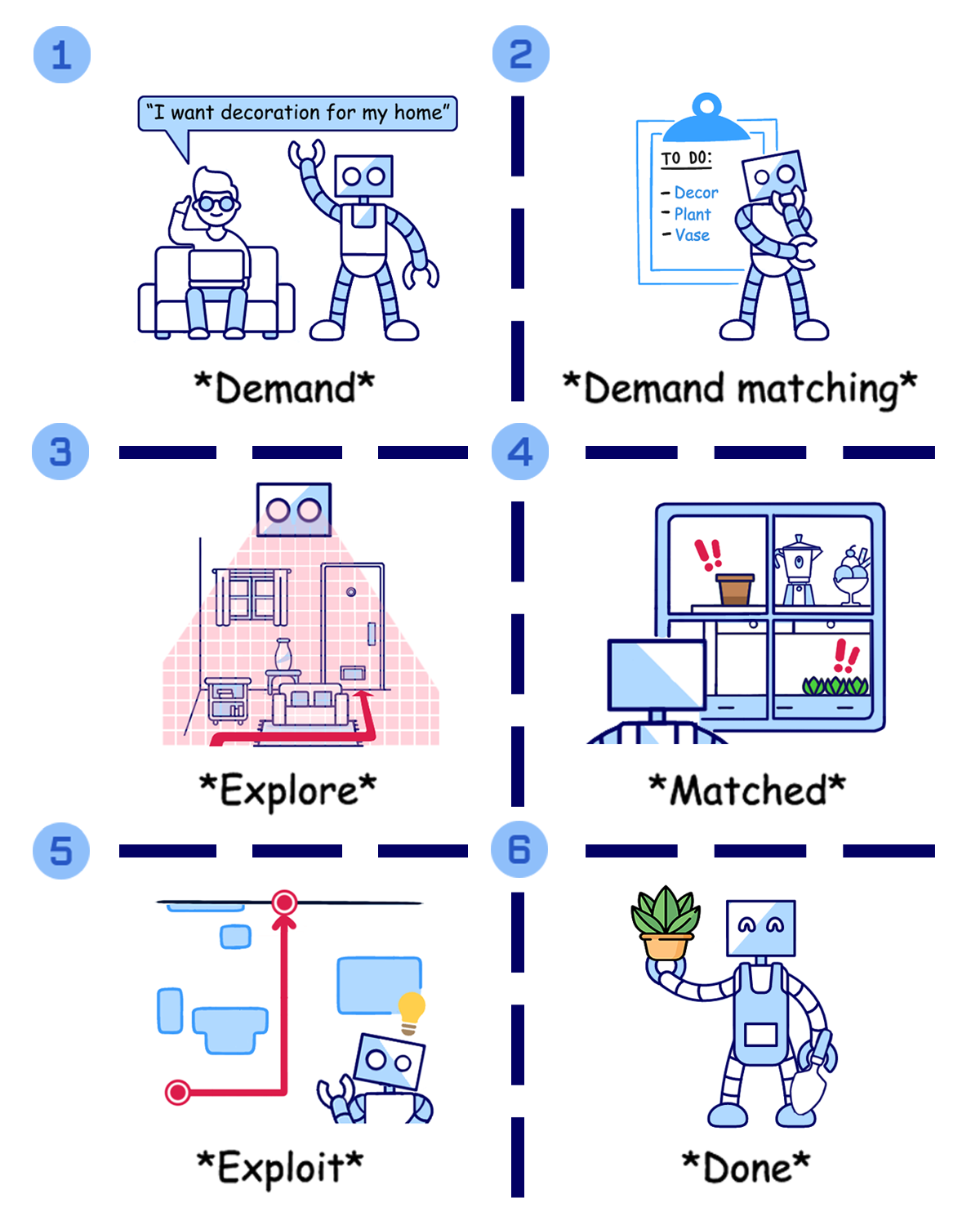}
  \caption{{\DDN} Framework for Demand-Driven Navigation Tasks. Upon receiving the instruction "I want decoration for my home", the agent continuously processes real-time observation perspectives and object information inputs. The agent continues this process until it identifies a match to the instruction. At this point, it transitions to the exploitation phase and moves towards the target object.}
  \label{fig:pipeline}
  \Description{{\DDN} Framework for Demand-Driven Navigation Tasks.}
\end{figure}

This paper introduces {\DDN}, a novel VLM-based framework for DDN that is inspired by the human dual-process cognitive framework and designed for robust, closed-loop decision-making. A key differentiator of our approach is its emulation of human cognitive and learning mechanisms to solve complex search tasks. Unlike many contemporary VLM-based navigation systems~\cite{instructnav,bridging} that rely on supplementary data such as multiple camera views, depth information, or pre-built maps, our system uses only egocentric, front-facing visual input. This design choice intentionally aligns our agent's perception with natural human search behavior and adheres to the practical constraints of real-world robotic hardware. Consequently, our system addresses a more realistic and challenging navigation task by mirroring the perceptual limitations and sequential decision-making inherent in human object search.

Analogous to human cognitive processes, {\DDN} incorporates a demand-matching module that leverages the semantic reasoning power of LLMs. This module translates ambiguous natural language instructions into a set of relevant, potential target objects, enabling the agent to identify and prioritize items that satisfy the user's underlying needs. Furthermore, we introduce a core decision-making system founded on dual-process theory~\cite{dual1,dual2,dual3}, which simulates human cognition via two distinct yet complementary components: a Heuristic Process (System-I) and an Analytic Process (System-II)~\cite{leapad,leapvad}. The Heuristic Process, akin to human intuition, facilitates rapid and efficient decisions based on prior knowledge. In contrast, the Analytic Process employs deliberate, step-by-step reasoning to refine these decisions, particularly in novel or challenging situations, and updates the knowledge base accordingly. Prior to navigation, we construct a knowledge base of high-quality, generalizable strategies. The Heuristic Process, continuously refined through supervised fine-tuning (SFT), draws upon this accumulated knowledge to guide initial actions. Meanwhile, the Analytic Process is invoked to update this knowledge base upon encountering obstacles or failures, creating a virtuous cycle of continuous learning and performance improvement. In contrast to direct end-to-end methods, our system integrates Chain of Thought (CoT)~\cite{cot} reasoning, enhancing the agent's ability to logically reason through complex scenarios and make transparent, informed decisions, rather than merely outputting final actions.

The main contributions of this paper are summarized as follows:
\begin{itemize}
\item We propose {\DDN}, a novel VLM-based framework for DDN tasks that simulates human-like cognitive mechanisms and learning processes to navigate based on implicit demands.
\item We introduce a dual-process decision-making module wherein the empirical Heuristic Process learns from the rational Analytic Process via a self-supervised mechanism, eliminating the need for human intervention.
\item {\DDN} uses the Analytic Process and a knowledge accumulation mechanism to progressively enrich a transferable knowledge base, enabling continual learning and effective generalization to novel navigation environments.
\item Extensive experiments on AI2Thor simulator~\cite{ai2thor} with the ProcThor dataset~\cite{procthor} demonstrate that {\DDN} outperforms existing single-view camera-only methods by 15\%.
\end{itemize}
\section{Related Works}

\subsection{Demand-driven Navigation}
Demand-driven Navigation (DDN)~\cite{ddn,mo-ddn} represents an advanced navigation paradigm where an agent must interpret and act upon high-level human demands expressed in natural language. This task diverges significantly from traditional visual navigation frameworks such as Visual Object Navigation (VON)~\cite{poni,zero,spatial,objectgoal,cliponwheels,learningobj}, which typically requires finding a specific object category (e.g., "find a chair"), and Visual Language Navigation (VLN)~\cite{voronav,safevln,navid,navgpt,navgpt2,generalscene,e2enav,bridging}, which relies on explicit, step-by-step instructions (e.g., "walk past the table and turn left"). In contrast, DDN centers on fulfilling an underlying user need (e.g., "I'm thirsty"), compelling the agent to infer which objects in the environment are relevant to satisfying the stated demand. This formulation makes DDN a more flexible and dynamic, shifting the focus from locating specific objects to addressing abstract human demands. The agent is thus required to perform a higher level of semantic interpretation and contextual inference based on the current visual scene, fostering a more generalized and human-aligned approach to navigation. Therefore, the primary challenge in DDN lies in this sophisticated process of interpreting abstract human needs and grounding them within a rich, dynamic visual context to guide action.

\subsection{Transitioning from Data-Driven to Cognition-Driven Navigation}
Early approaches to robotic navigation were predominantly data-driven, relying heavily on imitation learning~\cite{imitation1, imitation2} and reinforcement learning~\cite{reinforcement1,reinforcement2} to train reactive policies. These methods learn to map sensory inputs directly to actions by training on large-scale, often manually labeled, datasets. While such approaches have demonstrated promising results within their training distributions, they frequently exhibit brittleness and fail to generalize to out-of-distribution scenarios~\cite{CL-CoTNav,mcgpt}. This limitation arises primarily from a lack of deep semantic reasoning and contextual awareness, as the models often learn spurious correlations in the data rather than robust, causal relationships. Furthermore, these data-intensive methods typically incur high computational costs during extensive training and deployment~\cite{qqq, structured, dfrot}.

Motivated by the success of pre-trained vision-language models in a wide array of multi-modal reasoning tasks~\cite{licrocc, uniyolo, uist, learnact, guisurvey}, recent navigation research has shifted towards cognition-driven paradigms. This new wave of research adopts large-scale models as powerful cognitive priors, aiming to instill more robust and human-like navigation strategies~\cite{navigationvlmframework,WMNav,MapNav}. Inspired by human reasoning, these systems unify perception, action, and language to support flexible, language-driven goal specification~\cite{LOC-ZSON} and a deeper semantic understanding of the world~\cite{VLN-Game}. Unlike traditional methods that depend on structured instructions or explicit spatial maps, VLM-based approaches empower agents to interpret open-ended natural language commands and navigate effectively in complex, unfamiliar environments~\cite{VLNSurvey,SocialNav}. These modern systems emphasize multi-modal grounding, context-aware planning, and continual learning. To enhance transparency and robustness, they often incorporate explicit reasoning techniques, such as Chain-of-Thought (CoT) prompting, which encourages the model to generate a sequence of logical steps, thereby supporting more interpretable and generalizable navigation behavior.
\begin{figure*}[t]
  \centering
  \includegraphics[width=0.9\textwidth]{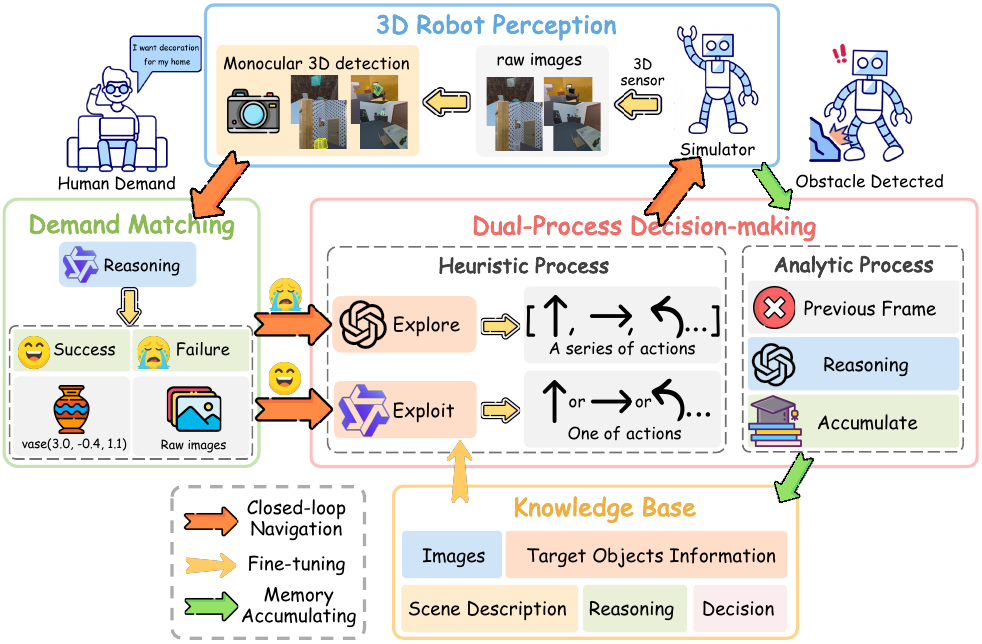}
    \caption{The detailed architecture of our proposed {\DDN}. The 3D Robot Perception module identifies objects based on the robot's perception of the environment. The robot then uses the detected objects and the human demand as input prompts for the demand matching module, which identifies the matched objects. This information enters the dual-process decision module, which drives scene description, reasoning, and decision-making. If no object matches the instruction, the Heuristic Process triggers the explore module to output a series of decisions for exploring unknown areas. Conversely, suppose the system finds a matching object. In that case, it activates the exploit module, refined through the knowledge base, to approach the target object. The Analytic Process analyzes the situation using a VLM whenever the system encounters obstacles, and it stores the corrected information as experience in the knowledge base.}
  \label{fig:pipeline}
  \Description{A detailed architecture diagram of the proposed {\DDN}.}
\end{figure*}

\section{Method}
In this section, we describe the design of {\DDN}, a closed-loop, anthropomorphic indoor navigation system. As shown in Figure ~\ref{fig:pipeline}, the system consists of five key components: 3D Robot Perception for identifying objects, an LLM for Demand Matching (Section ~\ref{sec:Demand Matching}), a Knowledge Base for storing experiences (Section ~\ref{sec:Knowledge Base}), and a dual-process decision-making module with the Heuristic Process (Section ~\ref{sec:Heuristic Process}) and the Analytic Process (Section ~\ref{sec:Analytic Process}). We begin by formally defining the DDN task in Section ~\ref{sec:Task Definition}, followed by detailed descriptions of each component.

\subsection{Task Definition}
\label{sec:Task Definition}
{\DDN} tasks require agents to locate the relevant objects in a given environment based on human demands and perceptual inputs. These tasks can be  formally described as a \textbf{Partially Observable Markov Decision Process (POMDP)}, defined as
\begin{equation}
    \mathcal{M} = (\mathcal{S}, \mathcal{O}, \mathcal{A}, \mathcal{E})
\end{equation}
where $\mathcal{S}$ represents the state space (the agent's current state), $\mathcal{O}$ is the observation space (including human demands and images), $\mathcal{A}$ is the action space (e.g., MoveAhead, RotateLeft, LookDown, etc.), and $\mathcal{E}$ is the action execution function. For instance, a user may state, "I am thirsty". In response, the agent must identify an object capable of quenching thirst based on the current visual observation and then execute the actions to locate the object. After each action, the agent receives updated observations and generates new actions based on the updated information to continue searching for the object. The task is considered complete when the distance between the agent and the target object is reduced to within 1.5 meters.

\subsection{Demand Matching}
\label{sec:Demand Matching}
A foundational principle of demand-driven navigation is that objects capable of fulfilling the same high-level human demand often share a set of key, underlying properties. For instance, paintings, houseplants, and statues are all suitable candidates for decorating a space. Although physically distinct, they all enhance visual appeal and complement the ambient environment. This intrinsic relationship between abstract human demands and object properties is grounded in commonsense or universal knowledge.

While LLMs are proficient at general reasoning based on instructions and properties, they can encounter difficulties when a perfect match for a given request is unavailable in the immediate environment. In such cases, standard LLMs may suggest suboptimal or functionally adjacent objects. For example, if a user requests, "I need something to hold my flowers", and no vase is present, the model might suggest a mug. Although a mug can technically hold water and flowers, it is a less-than-ideal choice domestically. This tendency to over-generalize based on basic affordances can significantly reduce the model's accuracy in selecting the most suitable object that aligns with user expectations. To mitigate this issue, we employ supervised fine-tuning (SFT) to further train the LLM, allowing it to better align object affordances with nuanced user requirements and contextual appropriateness. The fine-tuning enhances the model's capacity to identify and recommend the most fitting objects, ensuring it avoids suggesting less optimal options, even when an exact match is not readily apparent.

Formally, the Demand Matching module implements a target objects generation function $\mathcal{F : O \rightarrow \mathcal{P}, \mathcal{T}}$, where $\mathcal{O}$ represents the space of observation space including the human demands and visual information from images, $\mathcal{P}$ is the latent space of inferred demand properties, and $\mathcal{T}$ is the resulting space of candidate target objects. By leveraging a fine-tuned LLM for demand matching, the system benefits from an enhanced ability to reason over complex instructions and subtle object properties. This specialized capability prevents it from forcing matches between human demands and available objects when a direct correspondence is absent, ensuring more appropriate suggestions even in ambiguous scenarios. Please refer to Appendix ~\ref{appendix:anno_demand} for annotation details.

\begin{figure}[t]
  \centering
  \includegraphics[width=0.45\textwidth]{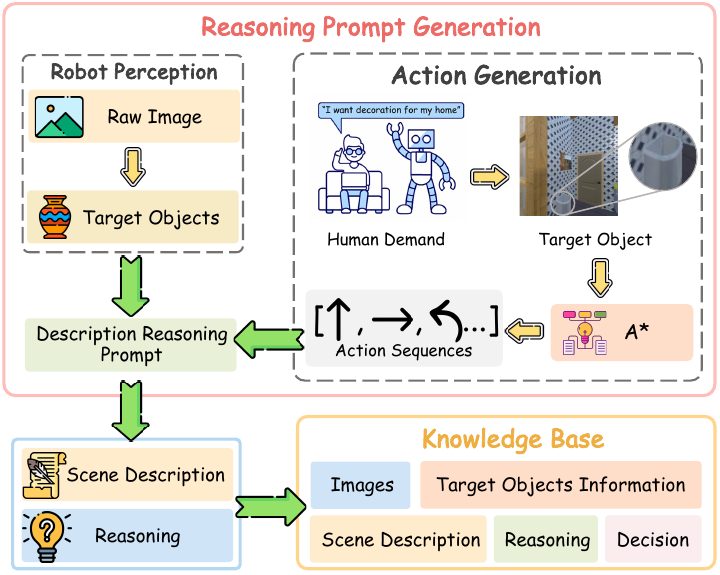}
  \caption{The Construction of the Knowledge Base. A random instruction is selected, and the target object is directly obtained from the simulator. The A* algorithm generates a trajectory consisting of actions based on the existing grid map and the target object's position. During the execution of these actions, the target object's position is determined using the current viewpoint image. When a matching object is detected, the VLMs generate a scene description and reasoning, which is added to the knowledge base.}
  \label{fig:knowledge_base}
  \Description{The Construction of the Knowledge Base.}
\end{figure}

\subsection{Knowledge Base}
\label{sec:Knowledge Base}
In indoor navigation scenarios, humans typically prioritize the approximate location of the target object and the scene description, focusing on navigable areas. This approach helps mitigate information overload, enhances reaction times, and reduces cognitive load. Consequently, we use scene descriptions to delineate passable areas, providing a sufficient contextual understanding of the environment. Additionally, Chain of Thought (CoT) reasoning enhances the model's inference process by leveraging the visual model's language-based reasoning capabilities. This approach contrasts with direct end-to-end methods by explicitly incorporating reasoning steps before generating final action decisions.

Inspired by LeapVAD~\cite{leapvad}, we generate additional data with outputs divided into three components: 1) \textbf{Scene Description}: This component outlines the environment based on the current viewpoint image, highlighting passable areas and task-related objects that may affect mobility. 2) \textbf{Reasoning}: This component leverages the visual-language model’s spatial awareness and commonsense reasoning capabilities based on the target’s location and the scene description to guide the agent’s navigation. 3) \textbf{Decision}: Drawing from the output of the reasoning component, the model generates the optimal decision for the current navigation task, ensuring the most effective action. This process incorporates knowledge transfer mechanisms to enhance navigation performance.
 
Building on this foundational approach of leveraging scene descriptions and reasoning for navigation optimization, we construct a knowledge base that integrates environmental data with dynamic decision-making processes. As illustrated in Figure ~\ref{fig:knowledge_base}, during the execution of the trajectory generated by the A* algorithm~\cite{astar}, once the target object is detected, the current viewpoint image $I$, target object $O_m$, and final action \textbf{S} are input into the VLMs to generate scene descriptions \textbf{D} and reasoning \textbf{R}, which are then recorded as an experience in the knowledge base. This accumulated experience can be progressively transferred to the Heuristic Process.  
As detailed in Section ~\ref{sec:Heuristic Process}, we utilize the collected data \{$O_m,I,\textbf{D},\textbf{R},\textbf{S}$\} to train the Heuristic Process's ability to follow instructions and adhere to formatting standards. Please refer to Figure \ref{fig:knowledge case} in Appendix \ref{appendix:anno_heuristic} for the data format.

\begin{figure}[t]
  \centering
  \includegraphics[width=0.47\textwidth]{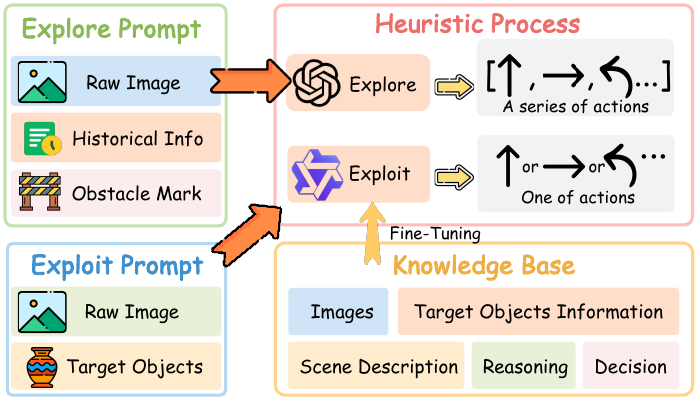}
  \caption{Detailed procedure of the Heuristic Process. When no target object is present, the Explore module directly utilizes the VLMs and the current viewpoint image, historical information, and obstacle mark to generate a series of actions for exploring unknown areas. In contrast, the Exploit module leverages a VLM fine-tuned with knowledge base information to output a single action that progressively moves towards the target object.}
  \label{fig:heuristic process}
  \Description{Detailed procedure of the heuristic process.}
\end{figure}

\subsection{Heuristic Process}
\label{sec:Heuristic Process} 
While the VLMs can effectively generate reasoning that aligns with decisions from path planning algorithms like A* algorithm~\cite{astar}, they often struggle to directly infer the causal relationship between reasoning and decisions. Moreover, their relatively slow processing speed can lead to redundancy and repetition, limiting their practical application in real-time indoor navigation. To address these challenges, we introduce the Heuristic Process in {\DDN}. This process is inspired by human navigation, where efficiency is gained through repeated practice and environmental familiarity, eventually allowing individuals to navigate with minimal cognitive effort. Formally, the Heuristic Process module implements an action execution function $\mathcal{E}:\mathcal{T} \times \mathcal{S} \times \mathcal{A} \rightarrow \mathcal{S}$, where $\mathcal{T}$ represents the space of target objects, $\mathcal{S}$ depicts the state space, and $\mathcal{A}$ is the action space, as presented in Table~\ref{tab:action space}. As demonstrated in Figure~\ref{fig:heuristic process}, it consists of two modules: \textbf{Explore} and \textbf{Exploit}, both leveraging Chain-of-Thought (CoT) reasoning to adapt to new environments and optimize learned navigation execution. The Heuristic Process is expressed as:

\begin{equation}
\label{heuristic_equation}
    \textbf{D}, \textbf{R}, \textbf{S} = 
    \begin{cases}
        \text{Explore}(I, S', X) & \text{if len}(O_{m}) = 0 \\
        \text{Exploit}(I, O_{m}) & \text{else}
    \end{cases}
\end{equation}

\textbf{Explore.} 
The Explore module is activated when the Demand Matching module fails to identify a suitable target object within the agent's current field of view. In this mode, the system defaults to a broad, information-seeking search of the environment. It generates exploratory actions by leveraging the VLM to interpret its visual perspective and reason about promising, unvisited areas. The objective is to efficiently survey the surroundings to uncover previously overlooked objects or navigational paths. As illustrated in Figure~\ref{heuristic_equation}, this process begins with the system generating a concise Scene Description \textbf{D} from the current viewpoint image $I$. Subsequently, it engages in explicit Reasoning \textbf{R} to formulate a strategy for exploration. Based on this reasoning, the system makes a Decision \textbf{S}, which comprises a sequence of actions designed to navigate toward unvisited regions. To prevent inefficient behavior, the module incorporates a short-term memory of prior actions and recent rotations $S'$ to minimize redundant movements. Furthermore, if an obstacle impedes progress, a flag $X$ is set to true, prompting the system to dynamically replan and generate new exploratory actions, which ensures continuous progress without getting stuck in loops or depleting resources.

\begin{table}[t]
    \small
    \centering
    \caption{\textbf{{\DDN} Action Space}}
    \begin{tabular}{@{}l>{\hspace{1cm}}p{6.5cm}@{}}
        \toprule
        \textbf{Action} & \textbf{Definition} \\
        \midrule
        MoveAhead & Move forward by $0.25$ meters. \\
        \midrule
        RotateLeft & Rotate the agent $30^\circ$ to the left. \\
        \midrule
        RotateRight & Rotate the agent $30^\circ$ to the right. \\
        \midrule
        LookUp & Tilt the agent's camera upward by $30^\circ$. \\
        \midrule
        LookDown & Tilt the agent's camera downward by $30^\circ$. \\
        \midrule
        Done & Indicate that the goal has been found. \\
        \bottomrule
    \end{tabular}
    \label{tab:action space}
\end{table}

\textbf{Exploit.} 
Once the Demand Matching module successfully identifies a target object, the system transitions from exploration to exploitation. The Exploit module shifts the agent's strategy from a broad, open-ended search to precise, goal-directed action. This phase leverages a version of the VLM that has been fine-tuned on prior successful experiences stored in the knowledge base, optimizing it for goal achievement. This fine-tuning enhances the model's ability to make accurate predictions and decisions based on subtle environmental cues, enabling the system to execute complex navigation tasks with high precision. Guided by a Chain-of-Thought(CoT) process, the system first generates a Scene Description \textbf{D} based on the current viewpoint image $I$ and the information about the target objects $O_m$. It then performs focused Reasoning \textbf{R} to infer the most direct course of action toward the target, given \textbf{D} and $O_m$. Finally, it produces the most optimal Decision \textbf{S}, a single, high-confidence action focused on progressing toward the target. By concentrating on a single, decisive action rather than a sequence of potential actions, the system improves its decisiveness and computational efficiency, ensuring rapid and effective goal attainment.

\subsection{Analytic Process}
\label{sec:Analytic Process}
\begin{figure}[t]
  \centering
  \includegraphics[width=0.45\textwidth]{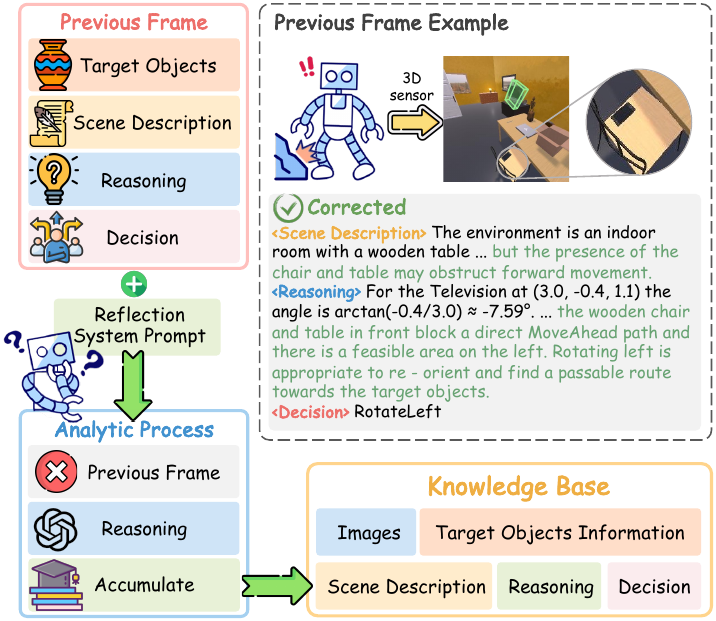}
  \caption{Detailed procedure of the Analytic Process. When the Heuristic Process encounters an obstruction, the Analytic Process intervenes by analyzing the previous frame to identify errors and generate corrected samples. These corrected samples are subsequently integrated into the knowledge base, supporting continuous learning.}
  \label{fig:reflect}
  \Description{Detailed procedure of the reflection mechanism.}
\end{figure}

The Analytic Process serves as the deliberative reasoning component of our framework, designed to reflect upon and learn from navigation failures, as illustrated in Figure~\ref{fig:reflect}. Through comprehensive pre-training on diverse, large-scale datasets, VLMs naturally accumulate a vast repository of world knowledge. This enables them to address complex challenges with sophisticated reasoning and insight, which is a capability that aligns perfectly with the objectives of the Analytic Process. This process moves beyond the rapid, intuitive responses of the Heuristic Process and instead relies on detailed causal analysis and deep contextual understanding to draw sound, generalizable inferences from navigational errors encountered in indoor environments.

Specifically, within the closed-loop operation of our system, any obstruction or failure encountered during the execution of the Heuristic Process triggers this reflective mechanism. When triggered, the Analytic Process utilizes the extensive world knowledge embedded within the VLMs to diagnose the situation. To do this, it ingests a snapshot of the state preceding the failure, including the target object's location, the scene description, and the specific reasoning and decision that led to the error. The system then carefully analyzes the root cause of the event, identifies the flaw in the previous logic, and generates a corrected chain of reasoning along with a revised decision. The valuable insights gained from this reflective process, a successful pairing of a problem state with a corrected solution, are subsequently integrated into the knowledge base. This integration establishes an iterative learning loop, enabling the system to learn from its failures continuously and progressively enhance its decision-making capabilities. This virtuous cycle ultimately leads to more informed and accurate navigation strategies for future tasks and directly improves the performance of the Heuristic Process over time.

\begin{table*}[t]
    \caption{\textbf{Performance comparison with single-view camera-only methods on DDN tasks.} \textbf{Seen Scene}: Train scene from ProcThor, \textbf{Unseen Scene}: Test scene from ProcThor, \textbf{Seen Ins}: Instructions used for training, \textbf{Unseen Ins}: Instructions used for testing, \textbf{NSR}: Navigation success rate, \textbf{SPL}: Navigation success rate weighted by the path length, \textbf{SSR}: Selection success rate.}
    \centering
    \tiny
    \setlength{\aboverulesep}{1pt}
    \setlength{\belowrulesep}{1pt}
    \setlength{\cmidrulewidth}{0.1pt}
    \renewcommand{\arraystretch}{1}
    \resizebox{1.0\textwidth}{!}{
    \begin{tabular}{lcccccccccccc}
    \toprule
    \multirow{3}{*}{\textbf{Method}} & \multicolumn{6}{c}{\textbf{Seen Scene}} & \multicolumn{6}{c}{\textbf{Unseen Scene}} \\ 
    \cmidrule(lr){2-7} 
    \cmidrule(lr){8-13} 
    & \multicolumn{3}{c}{\textbf{Seen Ins}} & \multicolumn{3}{c}{\textbf{Unseen Ins}} & \multicolumn{3}{c}{\textbf{Seen Ins}} & \multicolumn{3}{c}{\textbf{Unseen Ins}} \\
    \cmidrule(lr){2-4} 
    \cmidrule(lr){5-7} 
    \cmidrule(lr){8-10} 
    \cmidrule(lr){11-13} 
    & \textbf{NSR} & \textbf{SPL} & \multicolumn{1}{c}{\textbf{SSR}} & \textbf{NSR} & \textbf{SPL} & \multicolumn{1}{c}{\textbf{SSR}} & \textbf{NSR} & \textbf{SPL} & \multicolumn{1}{c}{\textbf{SSR}} & \textbf{NSR} & \textbf{SPL} & \textbf{SSR} \\
    \midrule
    Random~\cite{ddn} & 5.2 & 2.6 & 3.0 & 3.7 & 2.6 & 2.3 & 4.8 & 3.3 & 2.8 & 3.5 & 1.9 & 1.4 \\
    VTN-demand~\cite{ddn} & 6.3 & 4.2 & 3.2 & 5.2 & 3.1 & 2.8 & 5.0 & 3.2 & 2.8 & 6.6 & 4.0 & 3.3 \\
    VTN-CLIP-demand~\cite{ddn} & 12.0 & 5.1 & 5.7 & 10.7 & 3.5 & 5.0 & 10.0 & 3.6 & 4.0 & 9.3 & 3.9 & 4.0 \\
    VTN-GPT*~\cite{ddn} & 1.6 & 0.5 & 0 & 1.4 & 0.4 & 0.5 & 1.3 & 0.2 & 0.3 & 0.9 & 0.4 & 0.5 \\
    ZSON-demand~\cite{ddn} & 4.2 & 2.7 & 1.9 & 4.6 & 3.1 & 2.0 & 4.1 & 2.9 & 1.2 & 3.5 & 2.4 & 1.1 \\
    ZSON-GPT~\cite{ddn} & 4.0 & 1.1 & 0.3 & 3.6 & 1.9 & 0.3 & 2.5 & 0.7 & 0.2 & 3.2 & 0.9 & 0.2 \\
    CLIP-Nav-MiniGPT-4~\cite{ddn} & 4.0 & 4.0 & 2.0 & 3.0 & 3.0 & 2.0 & 4.0 & 3.7 & 2.0 & 5.0 & 5.0 & 3.0 \\
    CLIP-Nav-GPT*~\cite{ddn} & 5.0 & 5.0 & 4.0 & 6.0 & 5.5 & 5.0 & 5.5 & 5.3 & 4.0 & 4.0 & 3.0 & 2.0 \\
    FBE-MiniGPT-4~\cite{ddn} & 3.5 & 3.0 & 2.2 & 3.5 & 3.2 & 2.0 & 3.5 & 3.5 & 2.0 & 4.0 & 4.0 & 3.5 \\
    FBE-GPT*~\cite{ddn} & 5.0 & 4.3 & 4.3 & 5.5 & 5.0 & 5.5 & 4.5 & 4.3 & 4.5 & 5.5 & 5.0 & 5.5 \\
    GPT-3-Prompt*~\cite{ddn} & 0.3 & 0.01 & 0 & 0.3 & 0.01 & 0 & 0.3 & 0.01 & 0 & 0.3 & 0.01 & 0 \\
    MiniGPT-4~\cite{ddn} & 2.9 & 2.0 & 2.5 & 2.9 & 2.0 & 2.5 & 2.9 & 2.0 & 2.5 & 2.9 & 2.0 & 2.5 \\
    DDN~\cite{ddn} & 21.5 & 9.8 & 7.5 & 19.3 & 9.4 & 4.5 & 14.2 & 6.4 & 5.7 & 16.1 & 8.4 & 6.0 \\
    \midrule
    \rowcolor{gray!20}
    \textbf{{\DDN}(ours)} & \textbf{38.3} & \textbf{17.2} & \textbf{29.8} & \textbf{37.5} & \textbf{18.0} & \textbf{28.6} & \textbf{33.3} & \textbf{16.4} & \textbf{27.1} & \textbf{34.5} & \textbf{17.1} & \textbf{27.5} \\
    \bottomrule
    \end{tabular}
    }
    \label{tab:baseline}
\end{table*}

\section{Experiments}
\subsection{Data preparation}
Our experiments are conducted in the AI2Thor~\cite{ai2thor} simulator using the ProcThor dataset~\cite{procthor}. Following the setup of DDN~\cite{ddn}, we evaluate our model on 600 scenes (200 from each of ProcThor's train, validation, and test splits), containing 109 object categories. We use the instruction sets provided by DDN~\cite{ddn}, comprising 1692 training, 241 validation, and 485 test instructions.

\textbf{Data for 3D Robot Perception.} 
For the 3D Robot Perception Module, we collected 55K frames from AI2THOR, split into 70\% training, 10\% validation, and 20\% testing sets. Each frame includes annotations for object category, 2D bounding box, and 3D box attributes. The 3D box information was obtained by extracting the minimal bounding box from the object's point cloud, which was generated using the simulator's depth and segmentation maps. For clarity, we provide details and methods about data in Appendix ~\ref{appendix:anno_unimode}.

\textbf{Data for Demand Matching.} 
For the Demand Matching Module, we created a dataset of 10.7K instruction-attribute pairs. The construction process involved using LLMs to generate common attributes for objects associated with a given instruction. This was structured into a QA format, where the question (Q) was an instruction and a set of candidate objects, and the answer (A) was the shared attributes and a list of correct objects. The dataset was split into 7.2K training, 1.4K validation, and 2.1K test samples. Additional details are shown in Appendix~\ref{appendix:anno_demand}.

\textbf{Data for Heuristic Process.} 
For the Heuristic Process, we generated a dataset of 72K expert trajectories in AI2THOR. An optimal path was computed for each instruction using the A* algorithm~\cite{astar}. Once the target object was detected in the agent's field of view, we used a VLM to generate a rationale for the path taken. This data was formatted into a VQA structure, where the question (Q) was the target's location and the answer (A) was the scene description, reasoning, and decision sequence. This dataset, used to populate our Knowledge Base, was split into 80\% for training and 20\% for validation. Additional details are provided in Appendix~\ref{appendix:anno_heuristic}.

\subsection{Implementation Details}
For the 3D Robot Perception module, we utilize UniMODE~\cite{unimode}. The model is trained for 30,000 iterations with a batch size of 64 on four RTX6000 GPUs, a process that takes approximately 10 hours. Input images are resized from 448 × 448 to 640 × 512 pixels. We use the AdamW optimizer with a learning rate of $1.2 \times 10^{-3}$, and a weight decay of $1.0 \times 10^{-2}$. A CosineAnnealing scheduler is used, with a warm-up factor of 0.3333 for the first 2000 iterations. To ensure stability, gradient clipping is applied at a threshold of 35.0.

For the Demand Matching, we conduct SFT on Qwen2-VL-7B~\cite{qwen2vl} with LoRA, training with a batch size of 1 for 5 epochs on 4 RTX6000 GPUs in 1 hour. Key hyperparameters include a learning rate of $1 \times 10^{-5}$, gradient accumulation of 8, and a cosine scheduler. LoRA is applied with a rank of 64 and alpha of 128, while freezing the ViT layers.

For the Heuristic Process, we conduct SFT on Qwen2-VL-7B for 5 epochs using knowledge base samples, which takes about 14 hours, with consistent hyperparameters from the Demand Matching.

\subsection{Evaluation in Closed-Loop Navigation}
We conduct closed-loop experiments in AI2Thor, a widely-used open-source simulator, to evaluate the performance of {\DDN} on 400 scenes using three metrics: Navigation Success Rate (NSR), Navigation Success Rate weighted by the path length (SPL)~\cite{spl}, and Selection Success Rate (SSR). Table~\ref{tab:baseline} compares our method with competitive methods offered by DDN~\cite{ddn} on the ProcThor. 

As shown in Table~\ref{tab:baseline}, {\DDN} outperforms all single-view camera-only methods. With the LLM’s reasoning capabilities, the Demand Matching module exhibits high generalization, with minimal performance gaps between seen and unseen instructions. The main gap between seen and unseen scenes stems from the 3D Robot Perception. Additionally, as shown in Figure~\ref{fig:first} (c), {\DDN} demonstrates a smaller relative drop between seen and unseen conditions, proving its strong generalization ability. Furthermore, as shown in Table~\ref{tab:sota methods}, we compare {\DDN} with SOTA methods InstructNav~\cite{instructnav} on unseen scenes and instructions. Remarkably, {\DDN} achieves comparable performance, even though InstructNav leverages depth maps as additional input.

\subsection{Ablation Study}
We conduct ablation studies on Exploit CoT, and Reflection in a closed-loop navigation system. The experiments were carried out on both unseen scenes and unseen instructions, demonstrating the generalization and continuous learning capabilities of our {\DDN}.

\textbf{Ablation Study of Exploit.} The Exploit module uses a fine-tuned VLM to generate precise, deliberate single-step actions critical for approaching the target. To assess the impact of fine-tuning, we replace the Exploit module with a vanilla GPT-4 model not adapted for navigation and evaluate performance in two settings: (i) generating one action per step and (ii) generating a full action sequence. As shown in Table~\ref{tab:ablation_of_exploit_and_cot}, this substitution causes a significant performance drop, underscoring the importance of task-specific fine-tuning for accurate, step-wise decisions in the Exploit phase.

\begin{table}[t]
    \caption{\textbf{Performance comparison with SOTA methods on DDN tasks.} \textbf{I}: RGB image, \textbf{D}: Depth map.}
    \footnotesize
    \setlength{\tabcolsep}{6.5pt}
    \label{tab:sota methods}
    \begin{tabular}{lcccc}
        \toprule
        \textbf{Method} & \textbf{Modality} & \textbf{Reflection} & \textbf{NSR} & \textbf{SPL} \\
        \midrule
        InstructNav~\cite{instructnav} & D+I & $\times$ & 30.0 & 14.2 \\
        \midrule
        DDN~\cite{ddn} & I & $\times$ & 16.1 & 8.4 \\
        {\DDN}(ours) & I & $\checkmark$ & \textbf{34.5} & \textbf{17.1} \\
        \bottomrule
    \end{tabular}
\end{table}

\begin{table}[t]
    \caption{\textbf{Ablation study of Exploit and CoT.}}
    \footnotesize
    \setlength{\tabcolsep}{5.5pt}
    \label{tab:ablation_of_exploit_and_cot}
    \begin{tabular}{lcccc}
        \toprule
        \textbf{Method} & \textbf{Single Action} & \textbf{NSR} & \textbf{SPL} & \textbf{SSR} \\
        \midrule
        {\DDN}(w/o Exploit) & $\checkmark$ & 24.2 & 13.9 & 19.2 \\
        {\DDN}(w/o Exploit) & $\times$ & 29.1 & 12.4 & 20.2 \\
        {\DDN}(w/o CoT) & $\checkmark$ & 27.3 & 11.5 & 21.1 \\
        {\DDN} & $\checkmark$ & \textbf{34.5} & \textbf{17.1} & \textbf{27.5}  \\
        \bottomrule
    \end{tabular}
\end{table}

\textbf{Ablation study of CoT.} In both the Heuristic and Analytic Processes, we use CoT to guide the VLM in generating high-quality decisions. To evaluate its effect, we remove CoT and generate decisions directly. As shown in Table~\ref{tab:ablation_of_exploit_and_cot}, this results in a significant performance drop, underscoring its importance within the dual-process framework. In the Heuristic Process, CoT helps leverage environmental information for decision-making. In the Analytic Process, CoT enables the VLM to analyze obstacles and uncover root causes, resulting in more accurate and informed decisions.

\begin{figure}[t]
  \centering
  \includegraphics[width=0.45\textwidth]{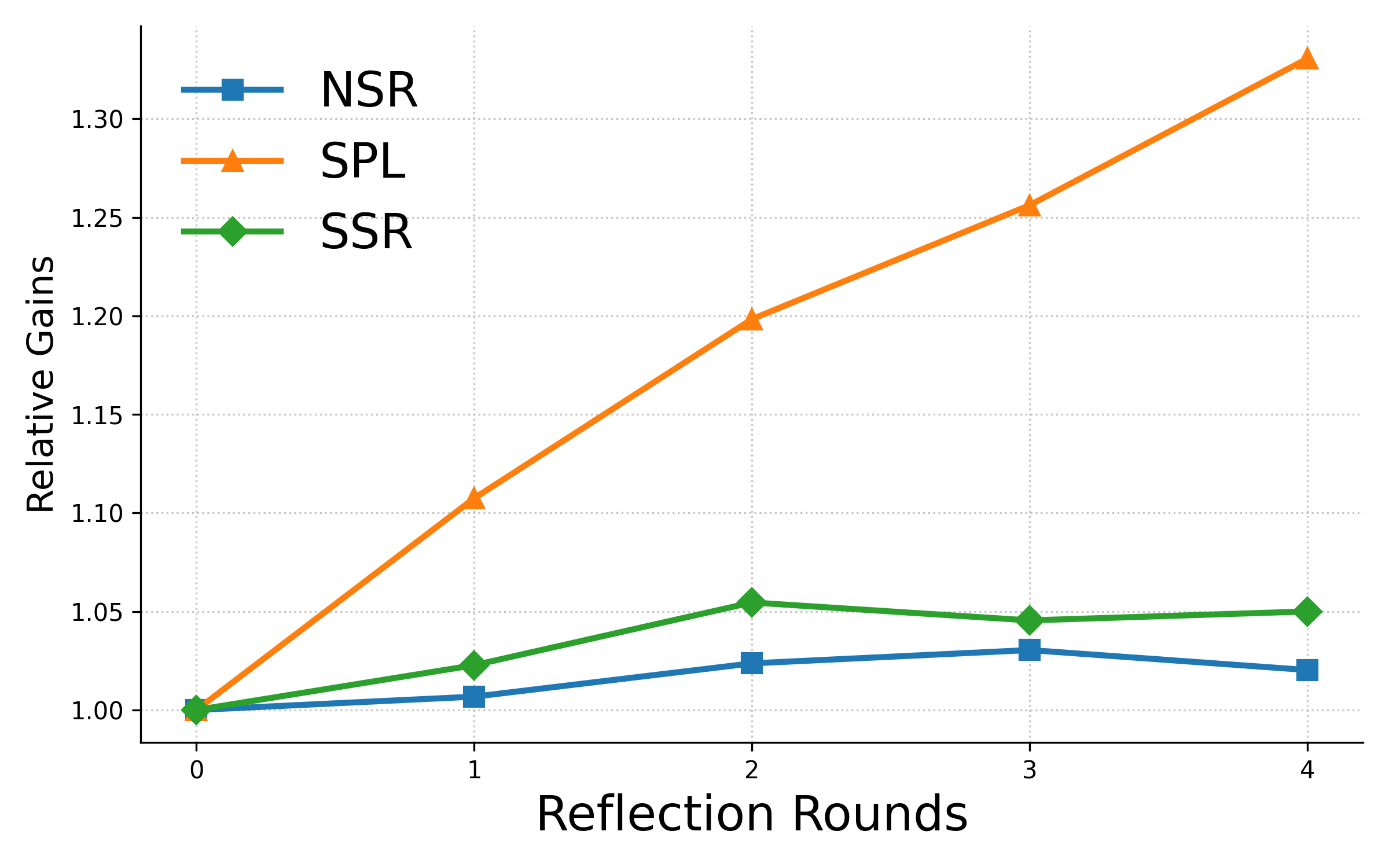}
  \caption{Ablation of Reflection. Four rounds of experiments were conducted, each consisting of 500 epochs. In each round, the experiences generated by the reflection mechanism were integrated into the knowledge base, and the VLM was fine-tuned accordingly.}
  \label{fig:ablation_of_reflection}
  \Description{Ablation of Reflection.}
\end{figure}

\textbf{Ablation study of Reflection.} Reflection is crucial for enhancing the continuous learning capabilities of our proposed {\DDN}. This mechanism promotes self-improvement by analyzing navigational failures and integrating corrected experiences into the knowledge base, allowing the agent to proactively accumulate knowledge in unseen environments. We conducted a closed-loop experiment over four iterative rounds of 500 epochs each to evaluate this capability. In each round, experiences generated by the reflection mechanism were used to augment the knowledge base and fine-tune the VLM. As shown in Figure~\ref{fig:ablation_of_reflection}, after four rounds of iterative learning, both SSR and NSR showed only marginal gains, which is an expected outcome as the reflection mechanism primarily enables the agent to recover from obstacles after they are encountered, leading to only modest improvements in these goal-completion metrics. In contrast, the SPL exhibited a significant increase. This substantial improvement in SPL demonstrates that as the agent accumulates experience, it learns to anticipate potential obstacles and take proactive detours, resulting in more efficient paths. This outcome underscores the effective continuous learning capability of {\DDN}, as its performance demonstrably improves with experience.

\section{Conclusion}
In this paper, we introduced {\DDN}, a novel dual-process closed-loop navigation system designed to emulate human attentional mechanisms. Central to our approach is a dual-process decision-making module that simulates human cognition, where a deliberative process is enhanced by Chain of Thought (CoT) reasoning to navigate complex scenarios. Furthermore, we integrated a cumulative knowledge base that enables the agent to continuously learn and self-improve from its experiences. Our comprehensive evaluations demonstrate that {\DDN} achieves state-of-the-art performance in closed-loop navigation tasks while requiring minimal training data, underscoring its efficiency and practical potential.

\textbf{Limitations and Broader Impacts.} 
The current {\DDN} system incorporates only short-term memory in the Explore module, lacking long-term memory to identify previously explored areas. Additionally, employing GPT-4 as a decision-making module in the Explore phase is computationally expensive and impractical for real-world deployment. Moreover, updating the Heuristic Process requires SFT whenever new experiences are added to the Knowledge Base, leading to inefficiency. In the future, we will design a new system focusing on long-term memory and end-to-end navigation.


\begin{acks}
This work is supported by the State Key Laboratory of Industrial Control Technology, China (Grant No. ICT2024A09).
\end{acks}

\bibliographystyle{ACM-Reference-Format}
\bibliography{main}

\clearpage
\appendix
\begin{algorithm}[t]
\caption{\DDN}
\label{alg:CogDDN}
\begin{algorithmic}
\STATE \textbf{Input:}
\STATE \quad Explore Prompt $P_e$
\STATE \quad Exploit Prompt $P_x$
\STATE \quad Reflection Prompt $P_r$
\STATE \quad Demand Prompt $P_m$
\STATE \quad Images \textbf{I} from agent
\STATE \quad Demand Instruction \textbf{M} from user

\STATE \textbf{Output:}
\STATE \quad Decisions $S$ for AI2THOR
\STATE

\STATE $O_a \leftarrow \text{3D\_Robot\_Detection}(\textbf{I})$
\STATE

\STATE DEMAND MATCHING (SECTION \ref{sec:Demand Matching})
\STATE $O_m \leftarrow \text{LLM}(P_m, O_a, \textbf{M})$
\STATE

\STATE HEURISTIC PROCESS (SECTION \ref{sec:Heuristic Process})
\FOR{$i = 1$ \TO length(max(sequence))}
    \IF{length($O_m$) = 0}
        \IF{length($S_q$) > 0}
            \STATE $S_i \leftarrow \text{POP}(S_q)$
        \ELSE
            \STATE $D_i,R_q,S_q \leftarrow \text{GPT}(P_e,\textbf{I},S_{q-1},X)$
            \STATE $S_i \leftarrow \text{POP}(S_q)$
        \ENDIF
    \ELSE
        \STATE clear($D_q, R_q, S_q$)
        \STATE $D_i,R_i,S_i \leftarrow \text{VLM}(P_x,\textbf{I},O_m)$
    \ENDIF
    \STATE send $S_i$
    \STATE $X \leftarrow \text{explore\_impede\_check}$
\ENDFOR
\STATE

\STATE ANALYTIC PROCESS (SECTION \ref{sec:Analytic Process})
\IF{detect\_hindrance() = True}
    \STATE $H \leftarrow \text{get\_hindrance\_info}()$
    \STATE $D_n, R_n, S_n \leftarrow \text{GPT}(P_r, H, \textbf{I})$ 
    \STATE $\textbf{B} \leftarrow \textbf{B} \cup \{ O_m, D_n, R_n, S_n \}$
    \STATE send $S_n$
\ENDIF
\end{algorithmic}
\end{algorithm}

\begin{figure}[t]
  \centering
  \includegraphics[width=0.47\textwidth]{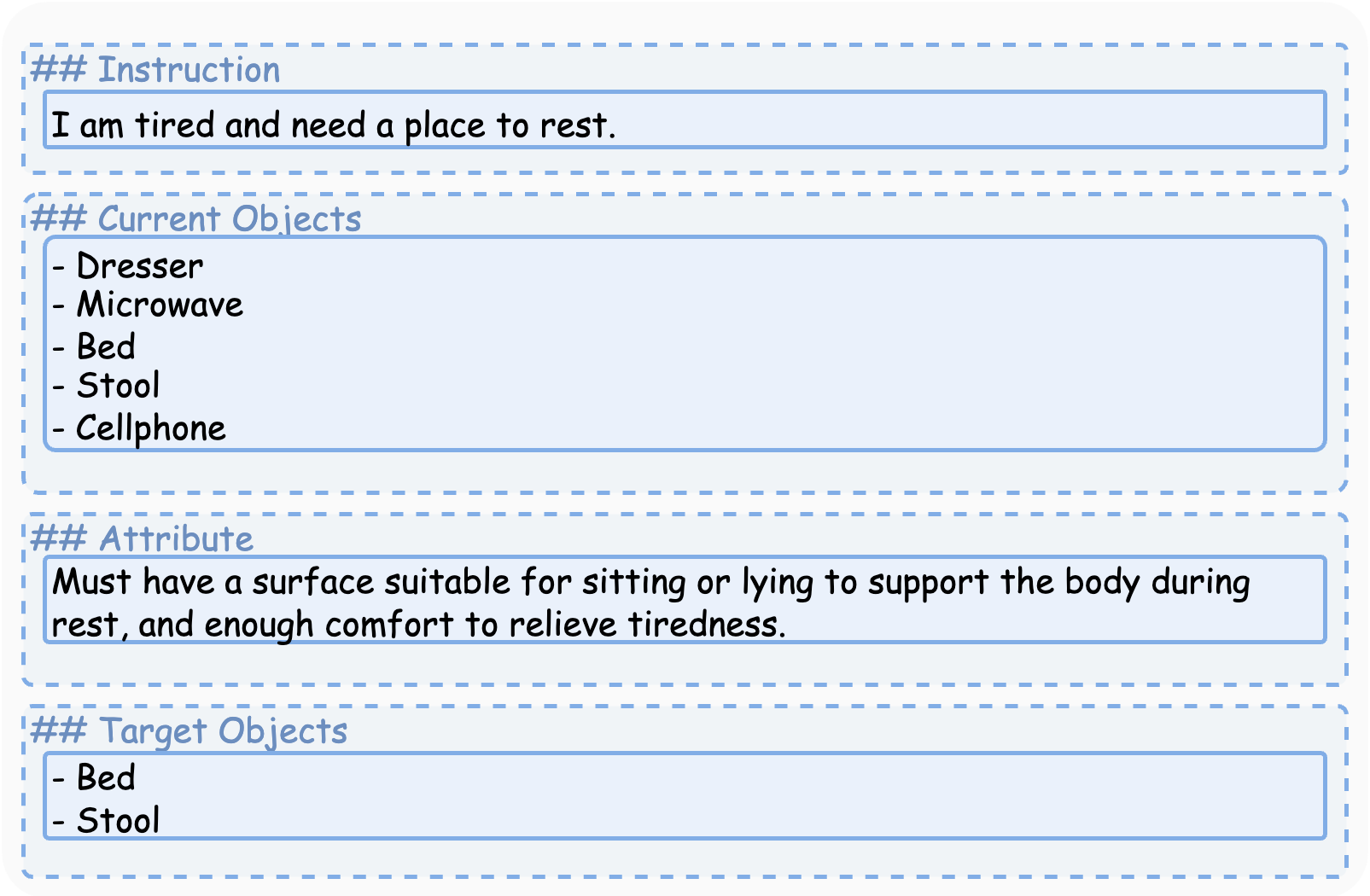}
  \caption{A case of Demand Matching dataset}
  \label{fig:demand case}
  \Description{A case of Demand Matching dataset}
\end{figure}

\section{Dataset Annotations}
In this section, we present examples of datasets utilized to train the 3D Robot Perception, Demand Matching and Heuristic Process.

\subsection{Annotations for 3D Robot Perception}
\label{appendix:anno_unimode}
The 3D Robot Perception Module was trained on data from the AI2THOR simulator. To derive the 3D bounding box ground truth, we first generated a point cloud for each object using the simulator's segmentation and depth maps. Subsequently, the minimal bounding box enclosing this point cloud was extracted to serve as the final, accurate 3D label.

Each frame in the dataset contains the following detailed annotations:

\begin{itemize}
    \item \textbf{bbox2D\_tight}: The 2D corners of the annotated tight bounding box, tightly fitting the object in 2D image space.
    \item \textbf{bbox2D\_proj}: The 2D corners projected from the 3D bounding box, representing the 2D location of the object from the camera's perspective.
    \item \textbf{bbox3D\_cam}: The 3D coordinates (in meters) of the eight corners of the bounding box in camera coordinates.
    \item \textbf{center\_cam}: The 3D coordinates of the object's center in camera space (in meters).
    \item \textbf{dimensions}: The 3D dimensions of the object in meters (width, height, and length).
    \item \textbf{R\_cam}: A 3x3 rotation matrix that represents the transformation between the object’s coordinate frame and the camera’s coordinate frame.
\end{itemize}


\subsection{Annotations for Demand Matching}
\label{appendix:anno_demand}
To train the Demand Matching Module, we constructed a specialized instruction-attribute dataset. This dataset was built from instruction-object pairs, where each instruction is associated with a set of candidate objects capable of fulfilling its requirements.

The construction process leverages a Large Language Model (LLM) to extract semantic features. For each instruction-object pair, the LLM is tasked with inferring the common attributes shared by the candidate objects that satisfy the instruction. These inferred attributes are then formulated into a Question-Answering (QA) format. The "question" component consists of the common attributes defining the target object, while the corresponding "answer" component contains a list of objects from the candidate set that match these attributes. A sample entry from the Demand Matching dataset is illustrated in Figure \ref{fig:demand case}.

\subsection{Annotations for Heuristic Process}
\label{appendix:anno_heuristic}
To generate a dataset for the supervised fine-tuning of our Heuristic Process, we utilized the AI2THOR simulator. Each scenario begins with a randomly selected instruction and a corresponding target object. We then compute an optimal trajectory to the target using the A* algorithm, which yields a discrete sequence of actions. As the agent executes this trajectory, it receives updated visual observations from the simulator at each step, enabling progressive localization of the target.

\begin{figure}[t]
  \centering
  \includegraphics[width=0.47\textwidth]{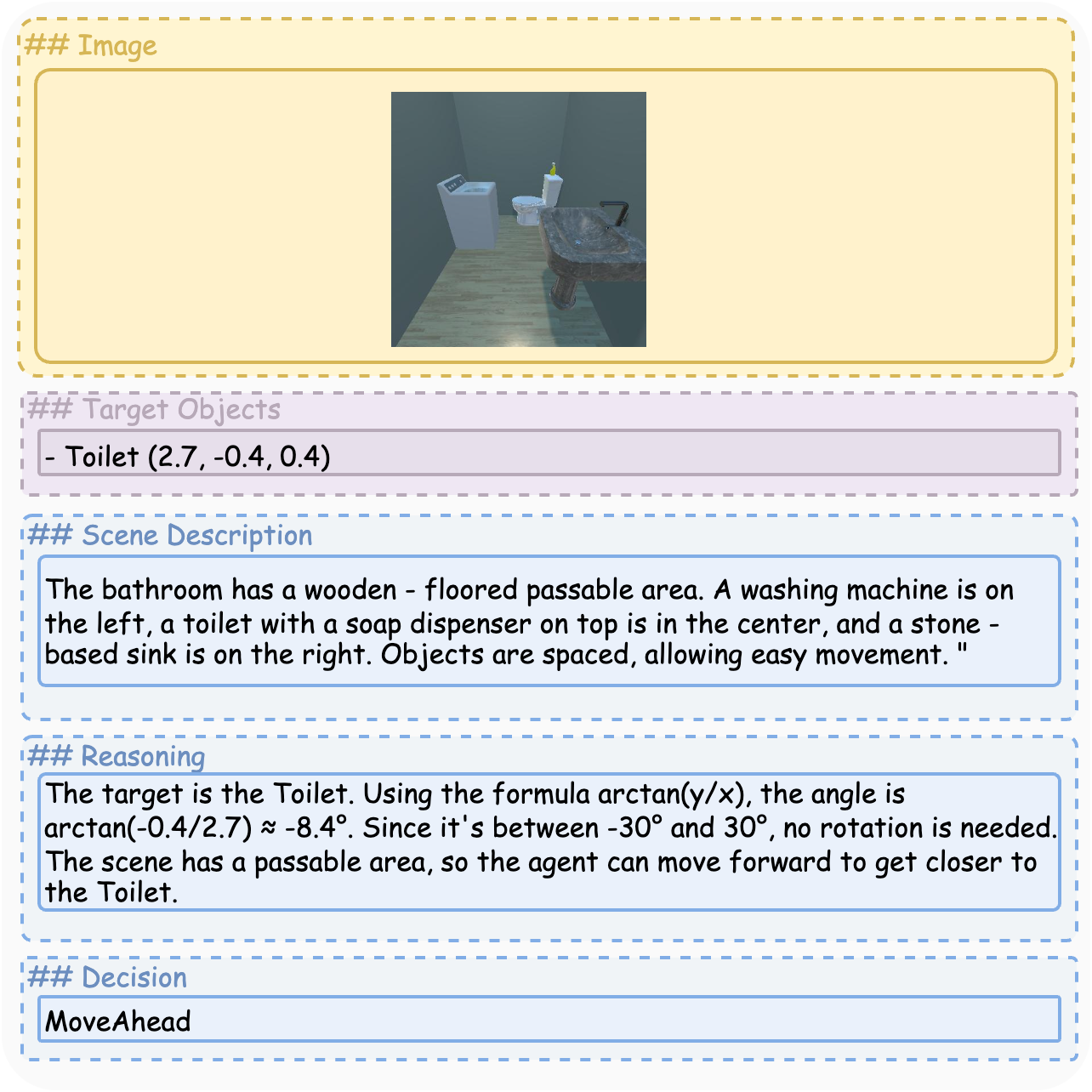}
  \caption{A case of knowledge base dataset}
  \label{fig:knowledge case}
  \Description{A case of knowledge base dataset}
\end{figure}

Upon successful detection of the object, a Vision-Language Model (VLM) is prompted to generate a structured output. This output includes a detailed scene description, a chain of reasoning that rationalizes the navigational choices, and the final decision. This information is then formatted into a Visual Question Answering (VQA) structure, as illustrated in Figure \ref{fig:knowledge case}. In this format, a query regarding the target object's status serves as the "question," while the VLM's generated text constitutes the "answer." These structured entries populate a knowledge base, providing the robust data required for our system to learn complex scene understanding and decision-making.

\section{Prompt Details}
\begin{figure}[t]
  \centering
  \includegraphics[width=0.47\textwidth]{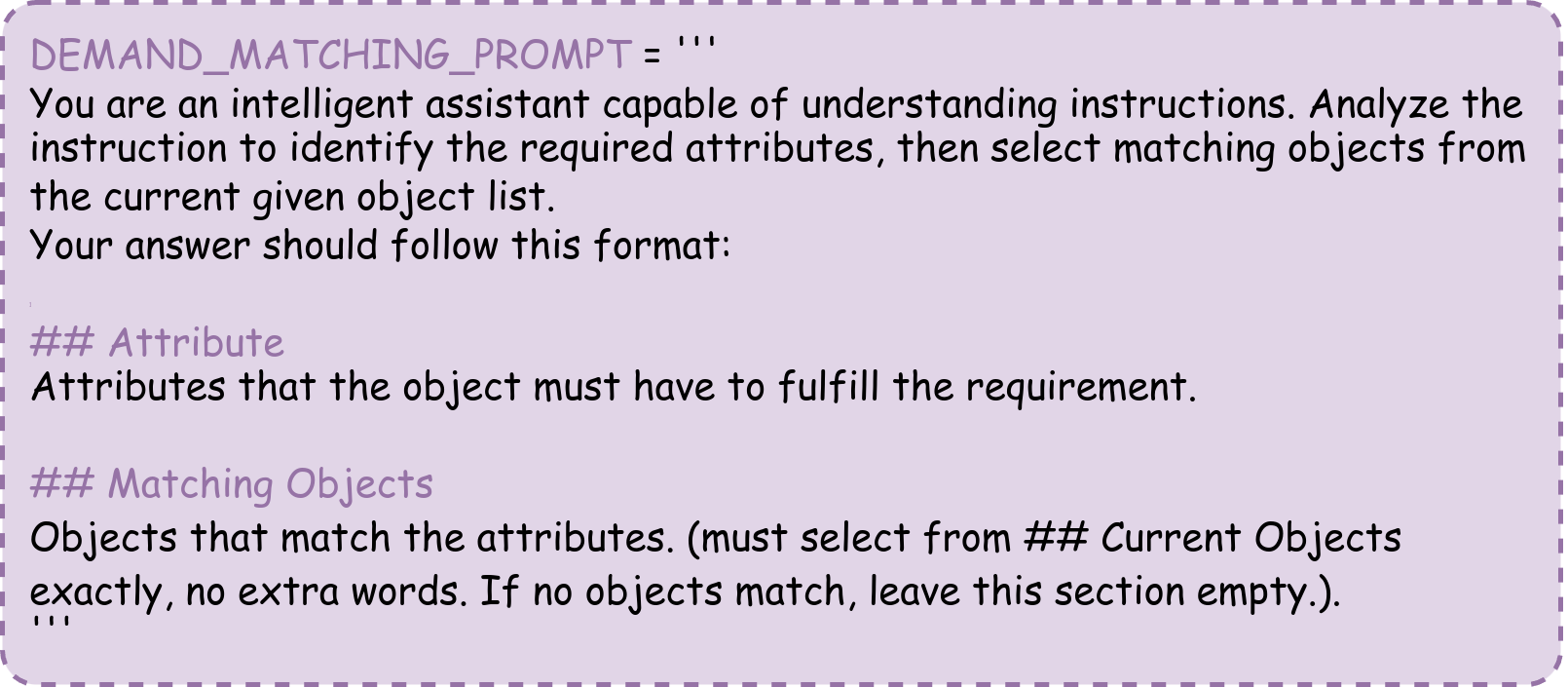}
  \caption{System prompt for Demand Matching}
  \label{fig:demand prompt}
  \Description{System prompt for Demand Matching}
\end{figure}

\begin{figure}[t]
  \centering
  \includegraphics[width=0.47\textwidth]{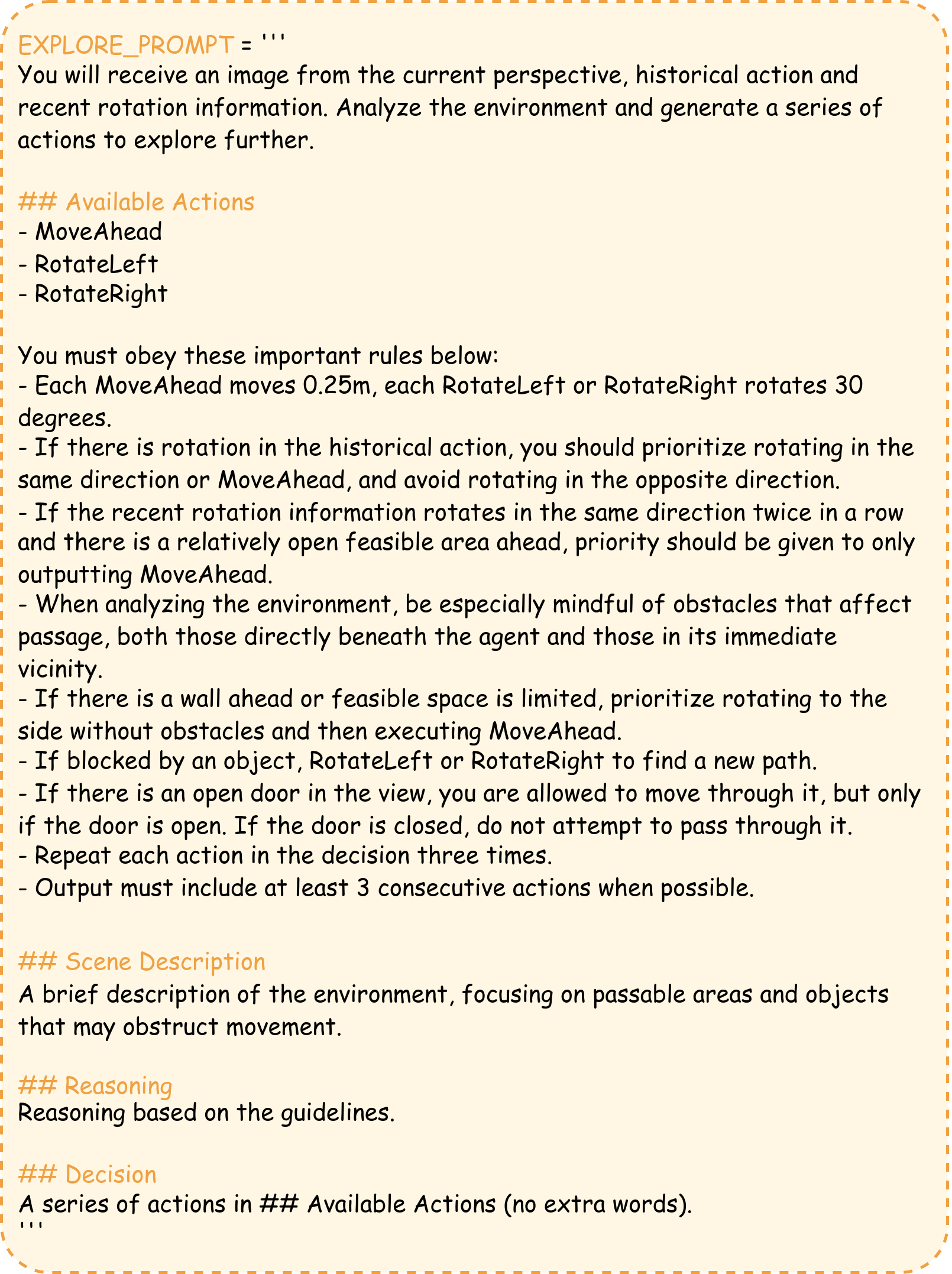}
  \caption{System prompt for Explore Module}
  \label{fig:explore prompt}
  \Description{System prompt for Explore Module}
\end{figure}

\begin{figure}[t]
  \centering
  \begin{minipage}{0.47\textwidth}
    \centering
    \includegraphics[width=\textwidth]{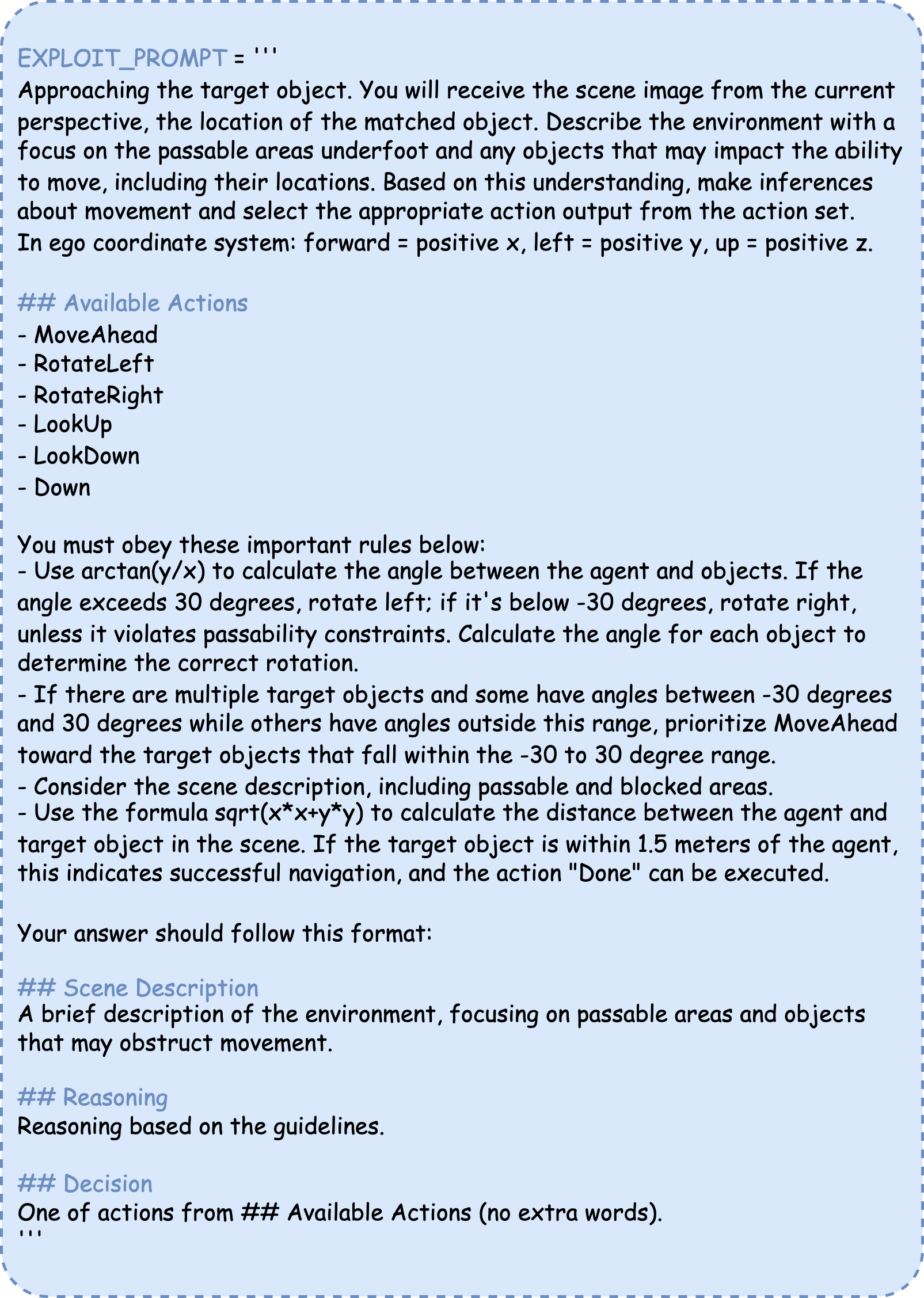}
    \caption{System prompt for Exploit Module}
    \label{fig:exploit prompt}
    \Description{System prompt for Exploit Module}
  
    \vspace{0.5cm} 
  
    \centering
    \includegraphics[width=\textwidth]{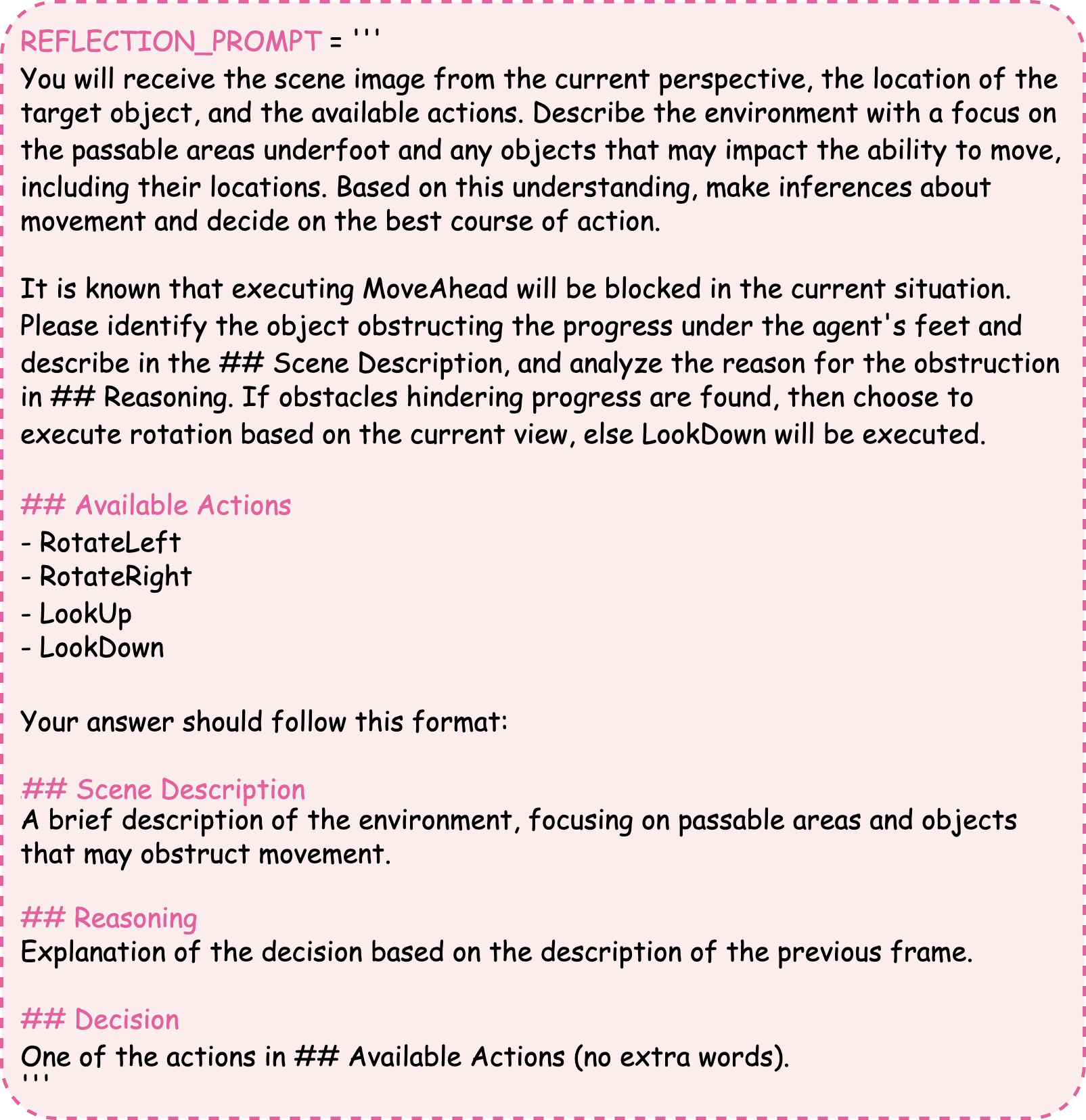}
    \caption{System prompt in the reflection mechanism}
    \label{fig:reflection prompt}
    \Description{System prompt for reflection mechanism}
  \end{minipage}
  
\end{figure}

\begin{figure}[t]
  \centering
  \begin{minipage}{0.47\textwidth}
    \centering
    \includegraphics[width=\textwidth]{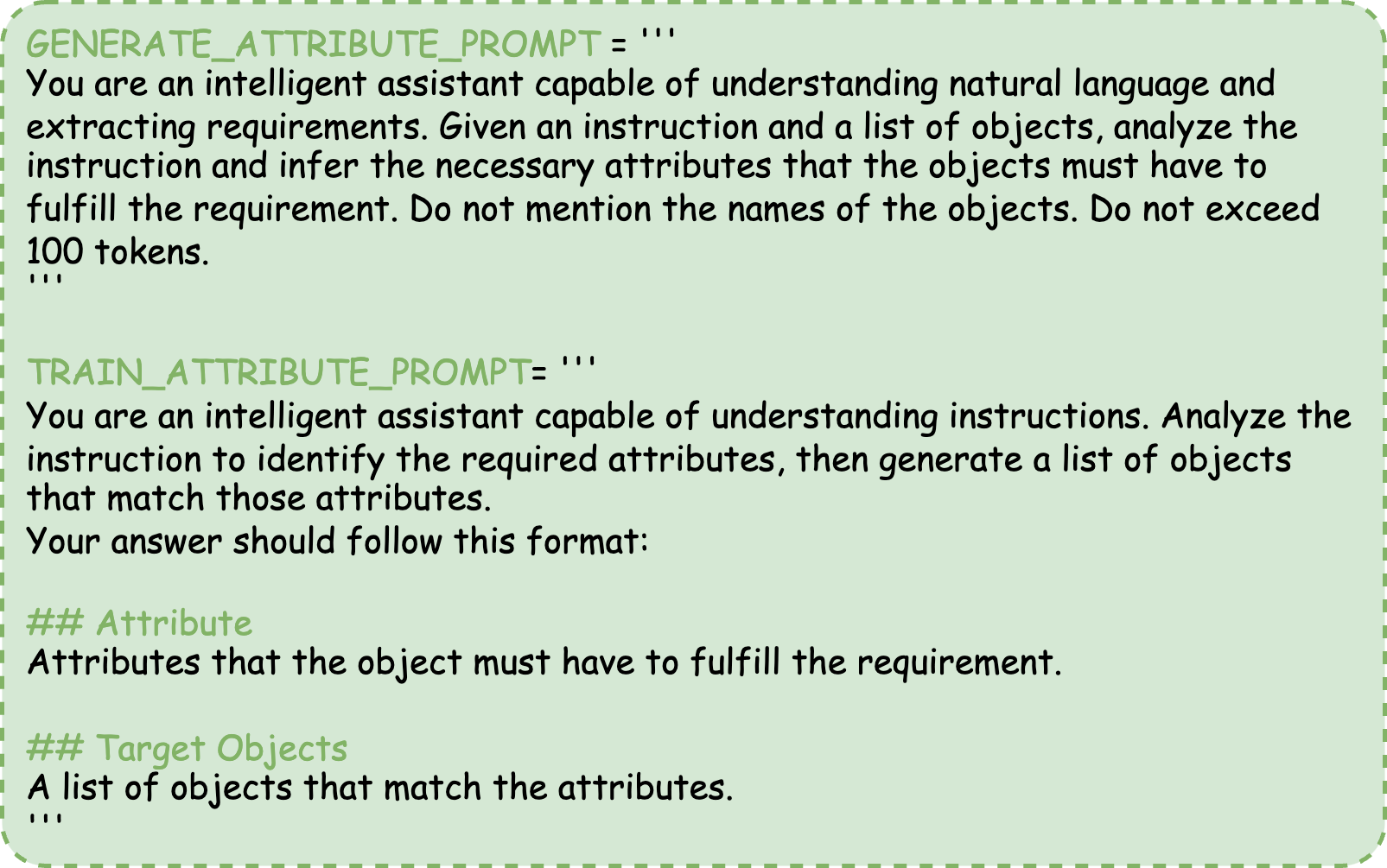}
    \caption{System prompt about the generation and train of attribute}
    \label{fig:attribute prompt}
    \Description{System prompt about the generation and train of attribute}
  
  \vspace{0.5cm} 
  
    \centering
    \includegraphics[width=\textwidth]{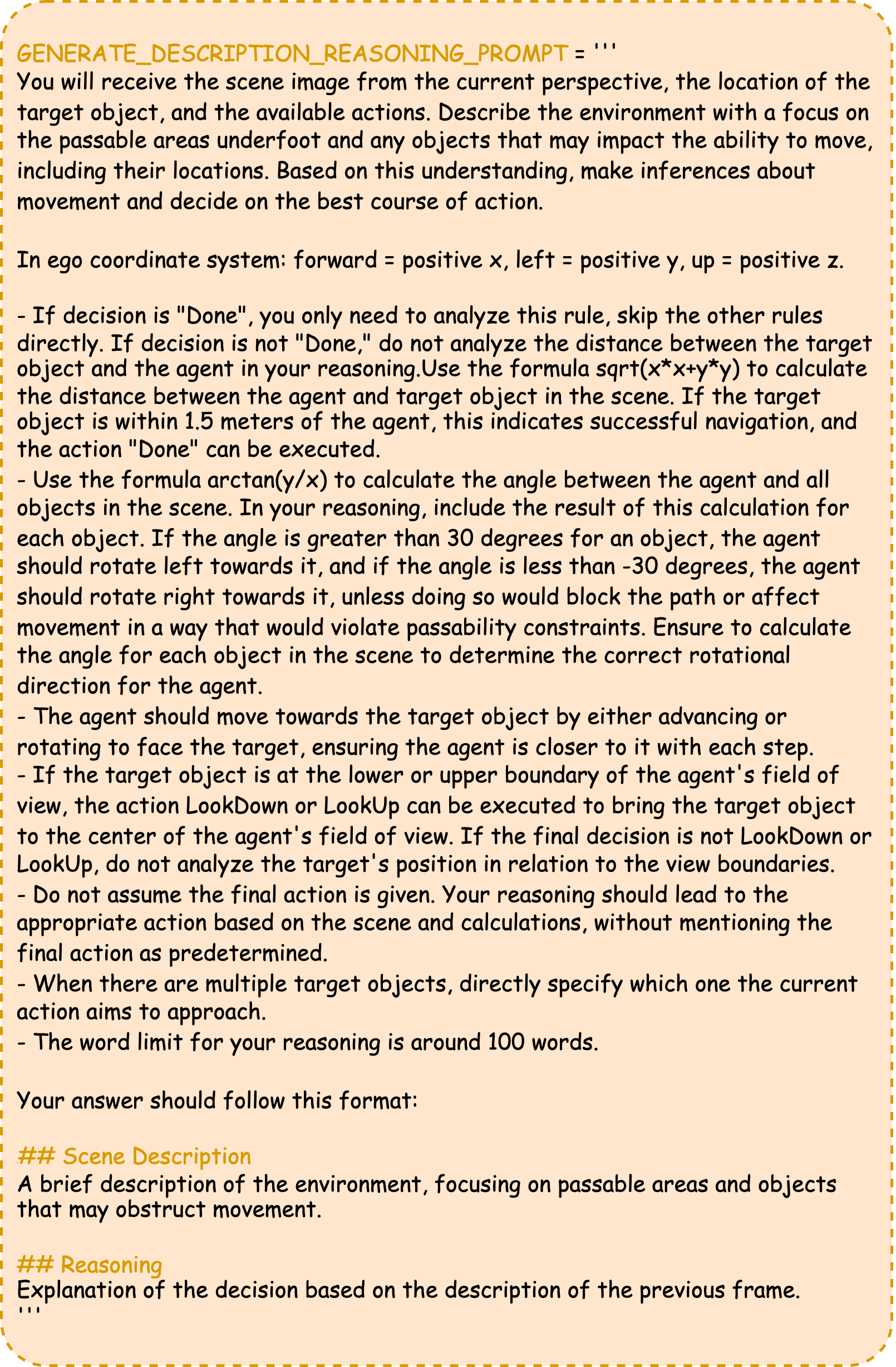}
    \caption{System prompt about the generation of scene description and reasoning}
    \label{fig:reasoning prompt}
    \Description{System prompt about the generation of scene description and reasoning}
  \end{minipage}
\end{figure}

We provide a detailed description of the demand matching prompt in Figure \ref{fig:demand prompt}, which is used to align the object in the field of view with the given instruction. Besides, Figure \ref{fig:explore prompt} illustrates the explore prompt employed by the Heuristic Process for exploration when no target object is present in the field of view. Additionally, Figure \ref{fig:exploit prompt} depicts the exploit prompt utilized by the Heuristic Process, which guides the approach toward the matched objects with precision and accuracy. Furthermore, Figure \ref{fig:reflection prompt} presents the reflection prompt, employed by the Analytic Process, which incorporates key elements from the previous prompts and includes criteria for detecting potential errors in the preceding frame. Moreover, Figure \ref{fig:attribute prompt} highlights the attitude generation and training prompt. Finally, as depicted in Figure \ref{fig:reasoning prompt}, the A* algorithm is employed to generate the Description and Reasoning prompts, which explain the selected path and the logical steps behind the decision-making process.

\clearpage

\clearpage

\begin{figure}[t]
  \centering
  \begin{minipage}{0.47\textwidth}
    \centering
    \includegraphics[width=\textwidth]{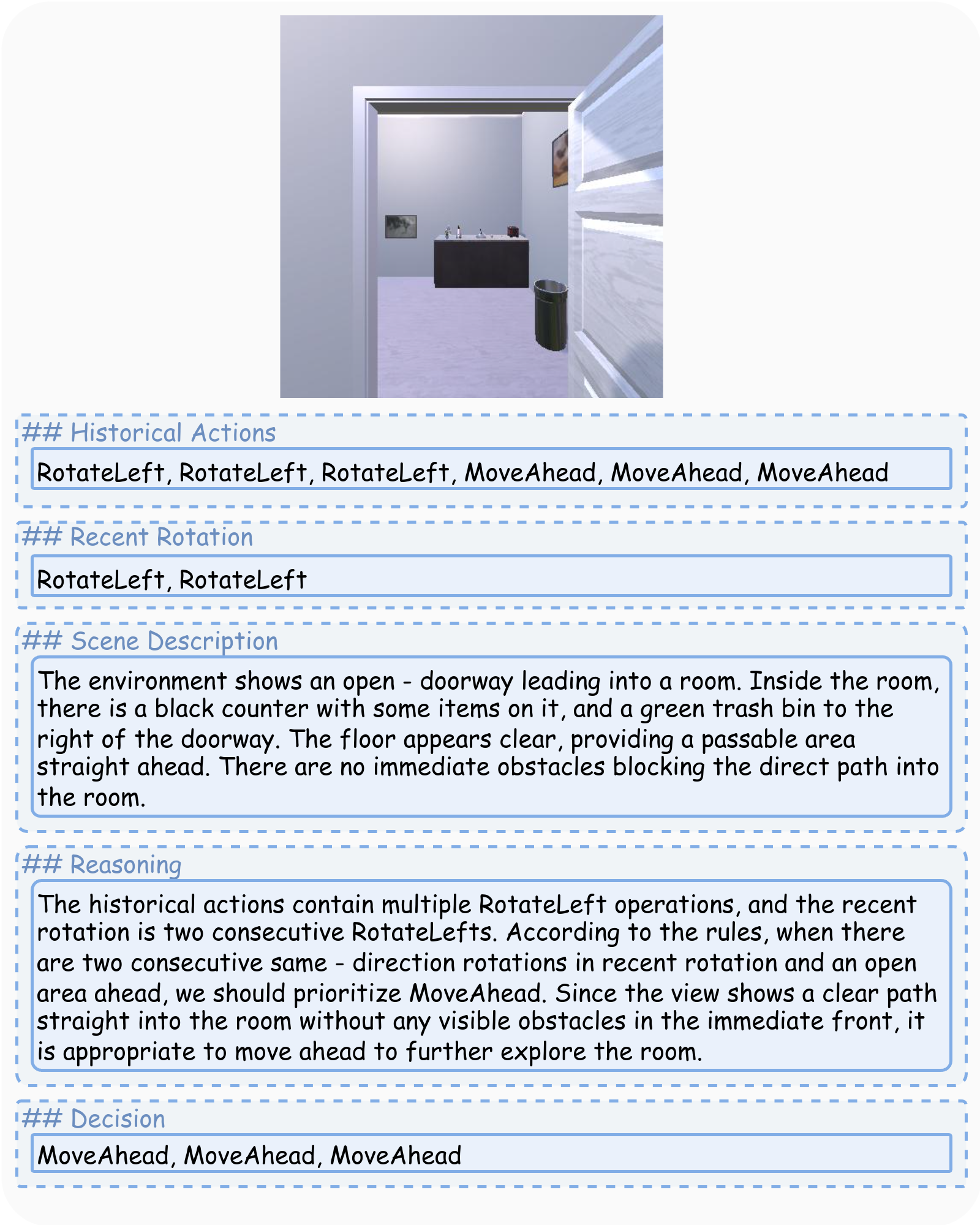}
    \caption{A case of Explore module}
    \label{fig:explore case}
    \Description{A case of Explore module}
  
  \vspace{0.4cm} 
  
    \centering
    \includegraphics[width=\textwidth]{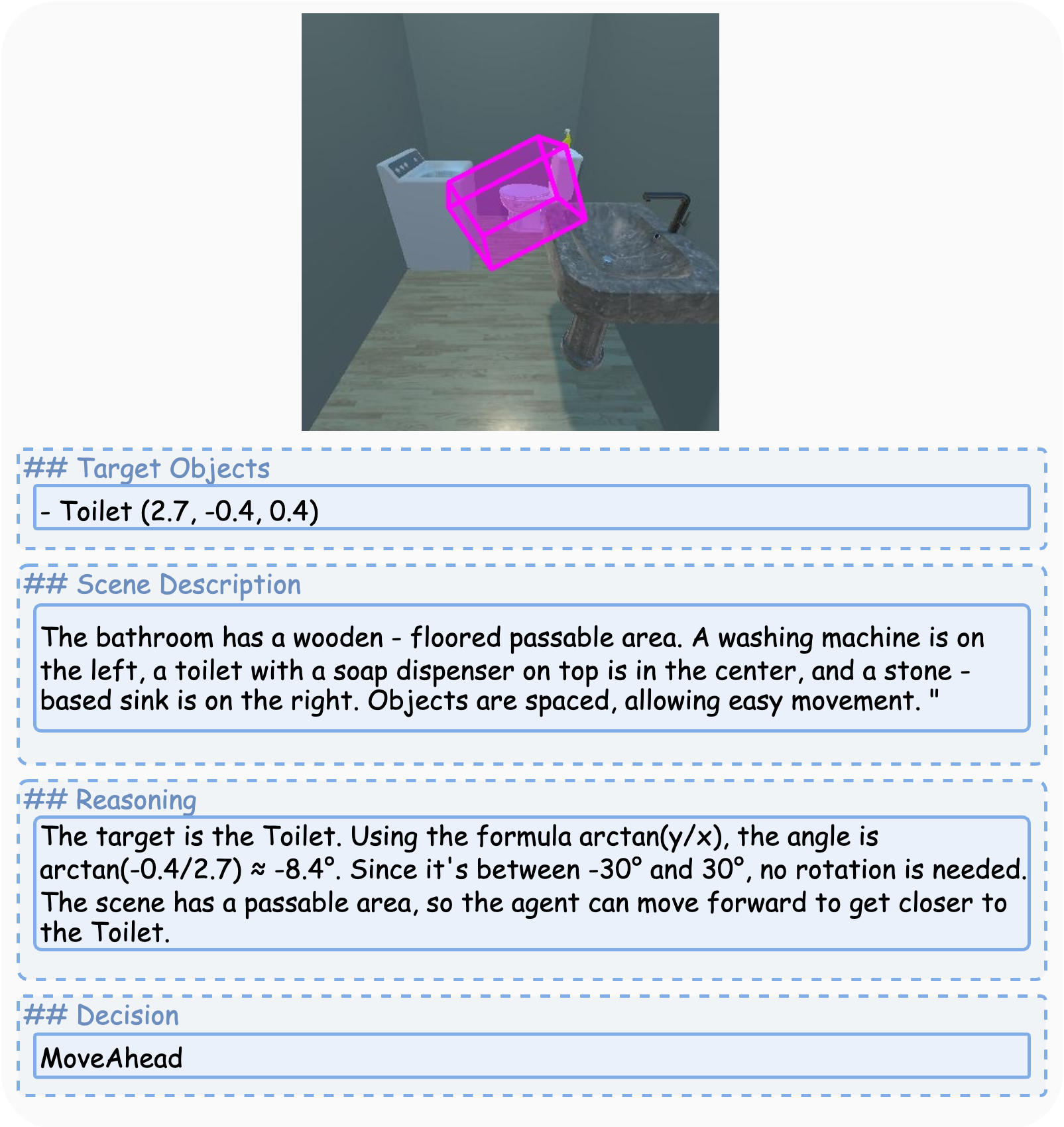}
    \caption{A straight case in an Exploit module}
    \label{fig:exploit case}
    \Description{A straight case in an Exploit module}
  \end{minipage}
\end{figure}

\begin{figure}[t]
  \centering
  \includegraphics[width=0.47\textwidth]{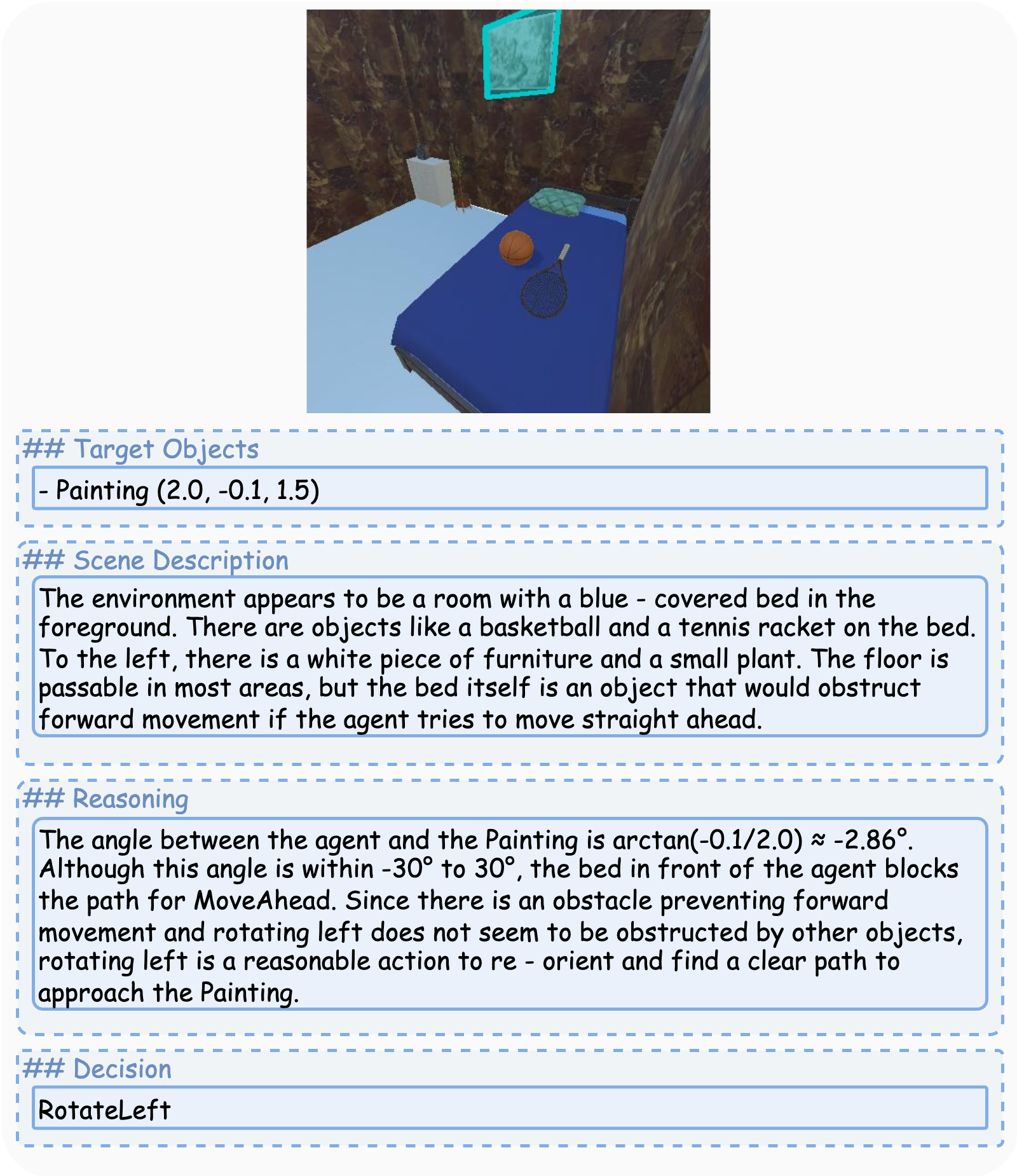}
  \caption{A bypass case in an Exploit module}
  \label{fig:exploit case2}
  \Description{A bypass case in an Exploit module}
\end{figure}

\begin{figure*}[t]
  \centering
  \includegraphics[width=1\textwidth]{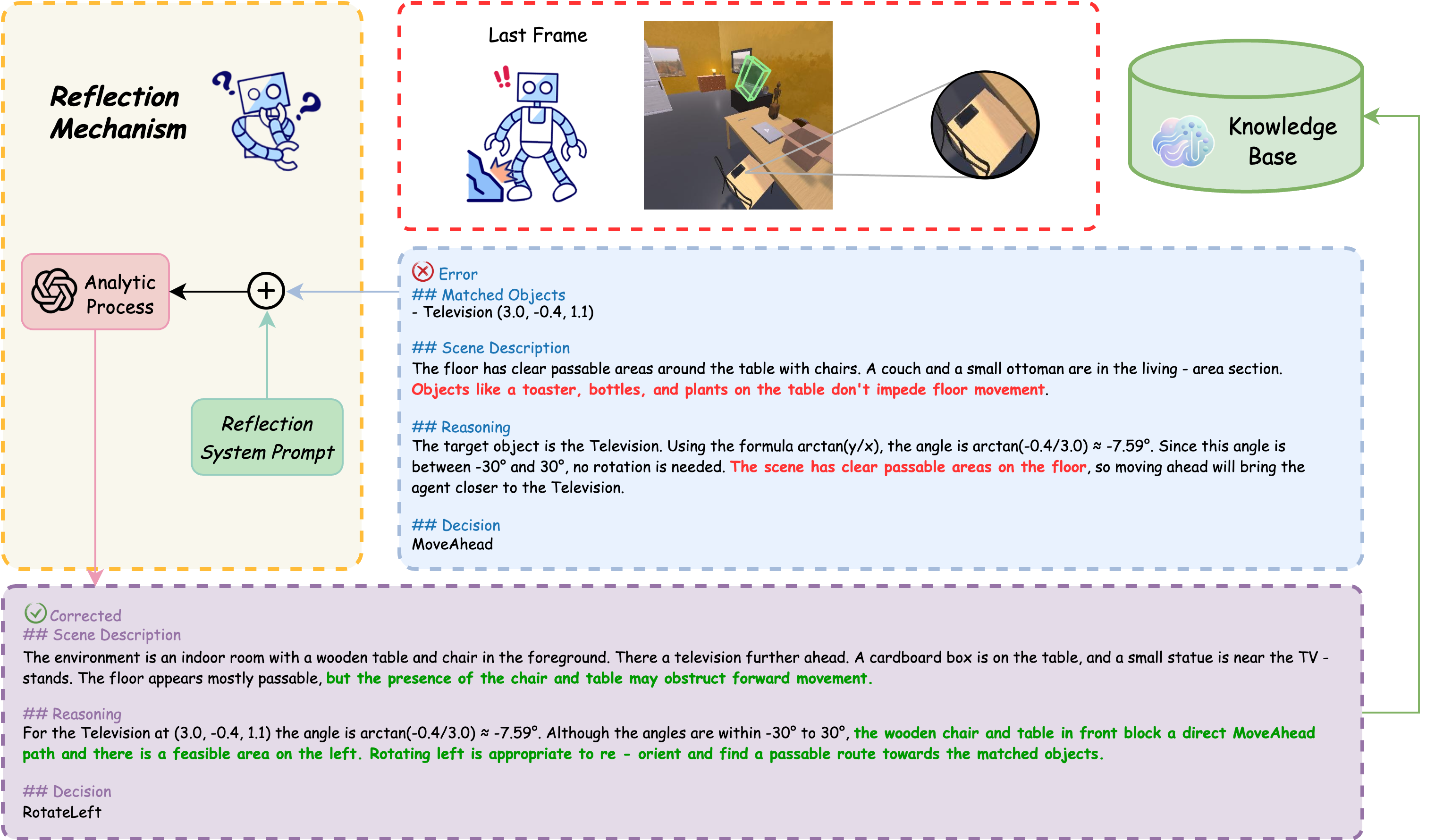}
  \caption{Case study for reflection mechanism.}
  \label{fig:reflect_case}
  \Description{Case study for reflection mechanism.}
\end{figure*}

\section{Visualization Cases}
\subsection{Cases of Explore}
Figure \ref{fig:explore case} presents a case to show the results using the Explore. When no object in the field of view matches the instruction, the VLMs take images of the current scene, historical actions, and recent rotations, generating scene descriptions, reasoning, and decisions, which include a series of actions to explore. As shown in this case, the historical actions included 'RotateLeft', and the recent rotation data revealed that the last two rotations were also 'RotateLeft' (consecutive rotations in one direction are recorded once). After passing through a door, a large open area was detected ahead. The VLM correctly decided on a series of 'MoveAhead' actions to explore forward while avoiding redundant exploration.

\subsection{Cases of Exploit} 
We also provide several examples to demonstrate the qualitative performance of our proposed {\DDN}, as illustrated in Figure \ref{fig:exploit case} and \ref{fig:exploit case2}. When an object matching the instruction is present in the field of view, the VLMs capture images of the current scene, generate scene descriptions and reasoning, and produce a decision that includes a single action to approach the target object. As shown in \ref{fig:exploit case}, the VLM correctly calculated the angle between the toilet and the agent, detected that there were no obstacles on the path to the toilet, and ultimately generated a 'MoveAhead' decision to approach it. Figure \ref{fig:exploit case2} presents a challenging scenario where the target object, a painting, is directly ahead, but a bed on the floor obstructs the agent's forward movement. The system effectively perceives the obstacle ahead, analyzes the environment, and promptly navigates around it towards a feasible area. In this case, with a wall on the right and a large feasible area on the left, the system correctly made a 'RotateLeft' decision, demonstrating its ability to effectively plan and adapt to the surroundings.

\subsection{Cases of reflection mechanism}
\label{appendix:case of reflect}
We also present an example in Figure \ref{fig:reflect_case} to illustrate the reflection mechanism. When the current path encounters obstacles, the image, description, reasoning, and decision are input into the Analytic Process to detect potential errors and provide reflective reasoning and decisions. As shown in this case, the initial analysis by the Heuristic Process incorrectly identifies the feasible area ahead, which requires immediate attention. In contrast, during the reflection phase, the Analytic Process accurately identifies the chair as the primary obstacle. It reassesses the previous 'MoveAhead' decision, suggesting that it may have been based on an inference error, as rotation would have been necessary to avoid colliding with the chair in this situation.

\end{document}